\documentclass[lettersize,journal]{IEEEtran}
\usepackage{amsmath,amsfonts}
\usepackage{algorithmic}
\usepackage{algorithm}
\usepackage{booktabs} 
\usepackage{array}
\usepackage[caption=false,font=normalsize,labelfont=rm,textfont=rm]{subfig}
\usepackage{textcomp}
\usepackage{stfloats}
\usepackage{float}
\usepackage[ colorlinks = true,
linkcolor = blue,
urlcolor  = blue,
citecolor = red,
anchorcolor = green,]{hyperref}

\usepackage{verbatim}
\usepackage{graphicx}
\usepackage{epstopdf}
\usepackage{cite}
\usepackage{adjustbox}
\usepackage{float}
\usepackage{multirow}
\usepackage{pifont}
\usepackage{amssymb}

\newcommand{\cyan}[1]{\textcolor{cyan}{#1}}

\usepackage[table,xcdraw]{xcolor}
\begin{document}
\setlength{\textfloatsep}{5pt}


\title{EarthMarker: A Visual Prompting Multi-modal Large Language Model for Remote Sensing}

\author{IEEE Publication Technology,~\IEEEmembership{Staff,~IEEE,}
\thanks{This paper was produced by the IEEE Publication Technology Group. They are in Piscataway, NJ.}
\thanks{Manuscript received April 19, 2021; revised August 16, 2021.}}

\markboth{}%
{Shell \MakeLowercase{\textit{et al.}}: A Sample Article Using IEEEtran.cls for IEEE Journals}

\author{
    Wei Zhang\textsuperscript{*}, Miaoxin Cai\textsuperscript{*},~\IEEEmembership{Graduate Student Member,~IEEE,} Tong Zhang,~\IEEEmembership{Graduate Student Member,~IEEE,} \\Yin Zhuang\textsuperscript{\dag}~\IEEEmembership{Member,~IEEE,} Jun Li,~\IEEEmembership{Fellow,~IEEE},  and Xuerui Mao\textsuperscript{\dag}
    
    \thanks{ {*} Wei Zhang and Miaoxin Cai contributed equally to this work.} \thanks{ {\dag} Co-corresponding author: Yin Zhuang and Xuerui Mao.}
    \thanks{Wei Zhang is with the Advanced Research Institute of Multidisciplinary Sciences, Beijing Institute of Technology, Beijing 100081, China, and also with the School of Mechatronical Engineering, Beijing Institute of Technology, Beijing 100081, China. (e-mail: w.w.zhanger@gmail.com)} 
    \thanks{Xuerui Mao is with the Advanced Research Institute of Multidisciplinary Sciences, Beijing Institute of Technology, Beijing 100081, China, and with the School of Mechatronical Engineering, Beijing Institute of Technology, Beijing 100081, China, and also with Yangtze Delta Region Academy of Beijing Institute of Technology, Jiaxing 314003, China. (e-mail: maoxuerui@sina.com).
    }
    \thanks{Yin Zhuang, Miaoxin Cai, and Tong Zhang are with the National Key Laboratory of Science and Technology on Space-Born Intelligent Information Processing, Beijing Institute of Technology, Beijing 100081, China. (e-mail: yzhuang@bit.edu.cn, 3120220667@bit.edu.cn, bit\_zhangtong@163.com).
    }
    \thanks{Jun Li is with the School of Computer Science and Hubei Key Laboratory of Intelligent Geo-Information Processing, China University of Geosciences, Wuhan, 430078, China (e-mail: lijuncug@cug.edu.cn).
    }
}
\maketitle

\begin{abstract}
Recent advances in prompt learning have allowed users to interact with artificial intelligence (AI) tools in multi-turn dialogue, enabling an interactive understanding of images. However, it is difficult and inefficient to deliver information in complicated remote sensing (RS) scenarios using plain language instructions alone, which would severely hinder deep comprehension of the latent content in imagery. Besides, existing prompting strategies in natural scenes are hard to apply to interpret the RS data due to significant domain differences. To address these challenges, the first visual prompting-based multi-modal large language model (MLLM) named EarthMarker is proposed in the RS domain. EarthMarker is capable of interpreting RS imagery at the image, region, and point levels by levering visual prompts (i.e., boxes and points). Specifically, a shared visual encoding method is developed to establish the spatial pattern interpretation relationships between the multi-scale representations of input images and various visual prompts. Subsequently, the mixed visual-spatial representations are associated with language instructions to construct joint prompts, enabling the interpretation of intricate content of RS imagery. Furthermore, to bridge the domain gap between natural and RS data, and effectively transfer domain-level knowledge from natural scenes to the RS domain, a cross-domain learning strategy is developed to facilitate the RS imagery understanding. In addition, to tackle the lack of RS visual prompting data, a dataset named RSVP featuring multi-modal multi-granularity visual prompts instruction-following is constructed. Extensive experiments are conducted to demonstrate the competitive performance of the EarthMarker. The proposed EarthMarker represents a significant advance in multi-granularity RS imagery interpretation under the visual prompting learning framework. Our code and dataset are available at \href{here}{\textit{https://github.com/wivizhang/EarthMarker}}.
\end{abstract}


\vspace{2cm}

\section{Introduction}

\IEEEPARstart{V}{isual} prompting is a current research hotspot in natural domains, aiming to interpret regions of interest in images with the guidance of visual marks (e.g., boxes, points, and scribbles)\cite{zhao2023chatspot,rezaei2024learning}.
{Particularly, injecting visual prompts into multi-modal large language models (MLLMs) can establish a joint prompting mechanism, which allows leveraging both visual and textual instructions. The joint prompting is a human-like expression way, enabling a more flexible and efficient interactive manner to understand images. Applying visual prompting to the remote sensing (RS) domain must account for the unique characteristics of RS imagery, which features scale variation, cross-category diversity, and complex contextual semantic information~\cite{zhang2024popeye,8113128}. Notably, a major challenge lies in effectively guiding models to recognize and interpret critical latent information from complicated RS imagery. However, it is difficult to efficiently describe the specific regions in complex RS images using plain language alone~\cite{chen2023position,kuckreja2024geochat}. Besides, existing MLLMs mainly achieve image-level visual-language alignment, which results in the restricted ability for detailed image understanding\cite{zhang2024rs5m}. Thus, it is indispensable to develop more concise and effective prompting mechanisms to enhance fine-grained visual reasoning in the RS domain.}

\begin{figure*}[!t]
	\centering
		\includegraphics[scale=0.17]{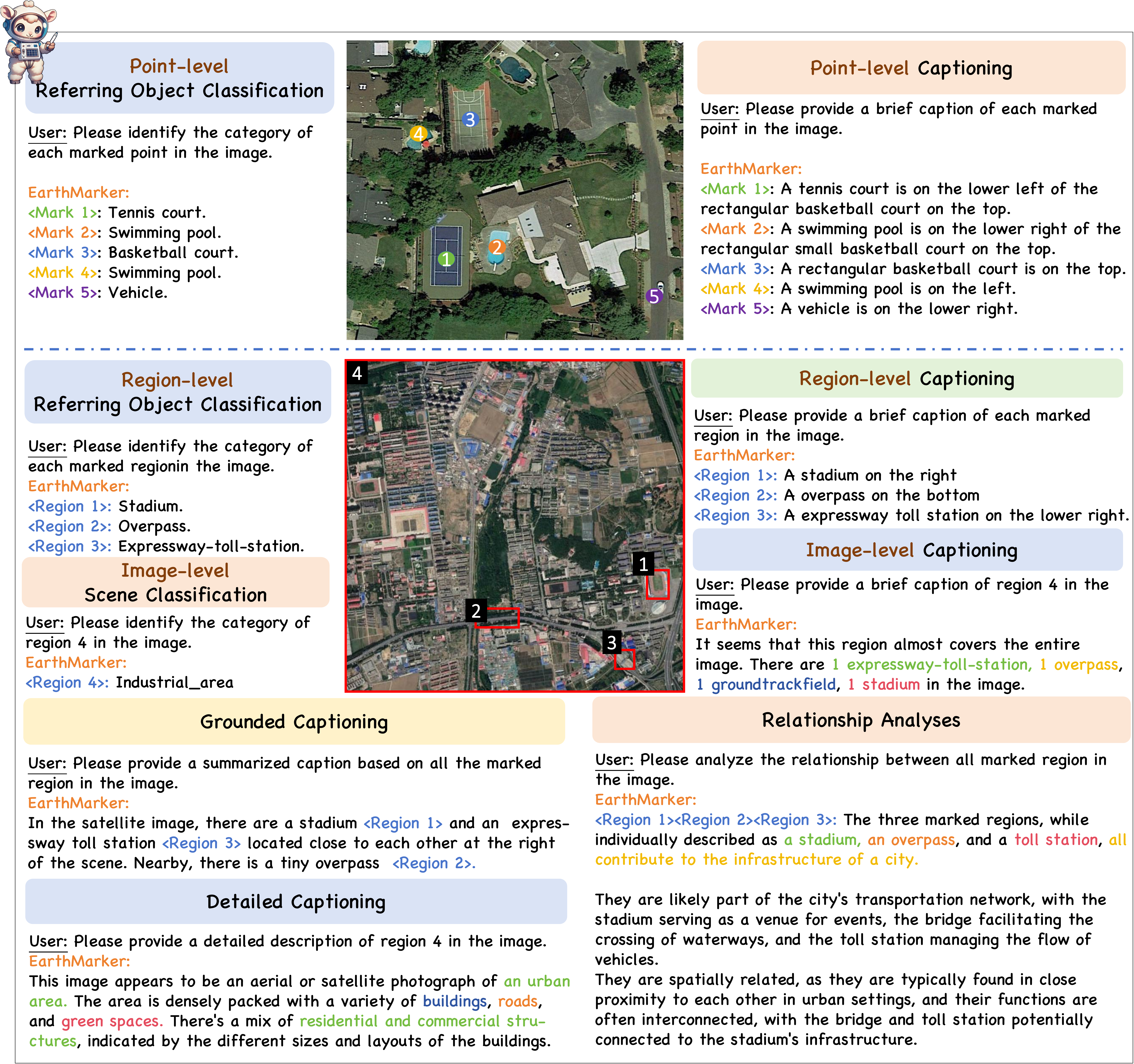}
	      \caption{Examples of multi-granularity (e.g., image-level, region-level, and point-level) RS imagery interpretation by the proposed EarthMarker, which excels in various visual tasks including scene classification, referring object classification, captioning, relationship analyses, etc.}
	\label{FIG:example}
\end{figure*}

Notably, prompting engineering~\cite{shin2020autoprompt,zhou2022learning} has been extensively studied in the natural language processing (NLP) community\cite{liu2023pre} and subsequently spread to the computer vision area. A key example is the Segment Anything (SAM)~\cite{kirillov2023segment} model, which utilizes multiple visual prompts to realize zero-shot segmentation adapted for various new image distributions.  GPT4RoI~\cite{zhang2023gpt4roi} and RegionBlip~\cite{zhou2023regionblip} have enabled MLLMs to complete region-level visual understanding tasks by training on region-text pairs. Osprey~\cite{yuan2023osprey} excels in pixel-level visual understanding but relies on pre-attached segmentation models, constraining its application range. Additionally, Ferret~\cite{you2023ferret} and SPHINX-V~\cite{lin2024draw} support free-shape visual prompting marks to achieve pixel-level image comprehension. Nonetheless, all these models are trained on natural scene data, leading to inferior performance when handling the RS imagery.

{In the RS field, there are limited works devoted to region-level fine-grained image understanding or visual prompting.} For example, RSVG~\cite{zhan2023rsvg} adopts language prompting to inquire and localize the specific object. Other representative RS MLLMs include EarthGPT~\cite{zhang2024earthgpt} and Geochat~\cite{kuckreja2024geochat}, which extend beyond image-level understanding to achieve region-level visual perception through training on visual grounding data. {In addition, SAM also shows limited effectiveness on RS imagery since its ViT backbone is pre-trained on large-scale close-range RS datasets\cite{osco2023segment,moghimi2024comparative}.} Inspired by SAM, RSPrompter~\cite{chen2024rsprompter} introduces an automated prompts generation to develop interactive segmentation tailored for RS data. {However, these RS MLLMs primarily realize image-text alignment and rely solely on language instructions, which struggle to uncover valuable information hidden within complex RS scenes and lack interactive flexibility. Furthermore, current visual prompting models including SAM and its variants are limited to image segmentation, lacking the ability for visual reasoning of complicated imagery. Therefore, the integration of visual prompting into MLLMs for detailed interpretation of RS data remains under-explored.}

To bridge the gap, a visual prompting MLLM named EarthMarker is proposed, leveraging visual prompts to extend the capability of MLLMs for region-level and point-level understanding in the RS domain for the first time. {As illustrated in Fig. \ref{FIG:example}, EarthMarker excels at the multi-granularity interpretation of RS imagery. Specifically, EarthMarker can switch levels to interpret an image, ranging from coarse-grained holistic scene-level to fine-grained region/object-level, and even point-level distinctions. Moreover, EarthMarker can complete a wide range of both coarse-grained and fine-grained RS visual tasks, including scene classification, referring object classification, image/region/point-level captioning, inter-relationship analyses, etc. Notably, the visual prompts are used to isolate specific areas and guide the model to interpret regional content in the entire RS image. Concretely, the visual prompts, i.e., bounding boxes and points, along with the RS images and the text instructions are provided as input to the large language model (LLM), adapt models toward specific predictions and tasks.} 

{In fact, there is a considerable gap between the different modalities involved in sparse visual prompts, dense RS image features, and textual features that operate on a semantic level. The significant differences in feature richness and structure make it challenging to align and integrate these modalities.} To address this problem, in our method, a sharing visual encoding method is developed. Specifically, the visual prompt is processed to RGB images analogously, which shares the same visual encoder with the input image. This strategy is beneficial for consistent feature extraction and understanding the relationship between visual prompts regions and the holistic image, enhancing the performance of the model under visual prompts learning. {The designed joint prompting mechanism, combining visual prompts and text instructions, enables EarthMarker to dynamically adapt to diverse tasks, interpretation granularity, and varying spatial resolution imagery.}

{The limited amount of RS data hinders achieving region and text mutual interaction, understanding spatial relationships, and object localization. To mitigate this limitation, general-domain data is utilized for mixed training. Most importantly, to enhance the \textit{visual prompts-image-text} alignment, the cross-domain learning strategy is proposed. }Specifically, in the first stage for multi-domain image-text alignment, EarthMarker is trained on the existing nature scene and RS caption data to obtain general image understanding and enhance the modeling of conceptual diversity. Subsequently, the model is further trained on the nature domain referring data to achieve spatial perception in images, beneficial for subsequent developing referring comprehension ability in the RS domain. {Lastly, in the RS visual prompting tuning stage, we leverage RS region-text and point-text instruction data, which features a variety of spatial resolutions, for domain adaption training. The proposed EarthMarker is equipped with holistic and point/region-level RS imagery interpretation capabilities}. Notably, the cross-domain learning training leverages the natural domain generalized knowledge and the RS domain expert knowledge for developing RS visual prompting MLLM. The multi-domain joint training is advantageous for enhancing the deep interpretation of fine-grained RS imagery and improving open-vocabulary reasoning capabilities. In addition, the updatable parameters of the model are disjoint, preventing interference between understanding images at different granularity and the capability to follow the visual prompt instruction. 

{Another challenge lies in existing visual prompting datasets ~\cite{you2023ferret,yuan2023osprey} are restricted to the natural scene, lacking RS semantics. Moreover, current public RS captions data tend to be overly simplistic and repetitive, without unique characteristics from the RS imagery.} It has become indispensable to construct a visual prompting dataset tailored to the RS domain. To this end, an RS visual prompting dataset named RSVP, featuring multi-modal large-scale visual-language joint instruction-following, is developed. In particular, diverse publicly available RS data are transformed and re-annotated into uniform conversation formats. Furthermore, part of the more high-quality caption data is generated from GPT4V\cite{achiam2023gpt}. Those captions are uniquely tailored with the distinctive characteristics of each RS imagery, thereby enhancing the richness and diversity of data. Through the data conversion and re-annotation from existing datasets and GPT4V, approximately 3.65 M image-point-text and image-region-text pairings are constructed. {In addition, RSVP includes images with a variety of spatial resolutions, covering a wide geographic distribution. These characteristics of the RSVP dataset equip EarthMarker to effectively handle both fine details and broader patterns, enabling it to generalize well in real-world applications.}

Extensive experiments are conducted on multi-type RS datasets to evaluate the performance of EarthMarker which is demonstrated to be superior to state-of-the-art (SOTA) specialist models, MLLMs, and visual prompting models in various RS visual tasks at different granularity. Specifically, for the zero-shot scene classification task, EarthMarker shows a significant improvement compared with other existing MLLMs. Notably, for referring object classification, EarthMarker achieves a Semantic Similarity (SS) score of 98.37 $\%$ using bounding boxes as visual prompts and 95.96 $\%$ using point prompts on DIOR-RSVG dataset\cite{10056343}. Furthermore, for image and region captioning tasks, EarthMarker also far exceeds other MLLMs and visual prompting models. In summary, the experimental results demonstrate that EarthMarker exhibits exceptional performance across a variety of multi-granularity RS image comprehension tasks and excellent zero-shot reasoning capability.

Our contributions can be summarized as follows.
\begin{itemize}
\item{\textit{First RS Visual Prompting MLLM, EarthMarker.}  
A visual prompting MLLM called EarthMarker is proposed in the RS domain for the first time. {EarthMarker can comprehend RS imagery under visual and text joint prompts, and flexibly switch interpretation levels, including image, region, and point levels. More importantly, the proposed EarthMarker fills the gap in visual prompting MLLMs for RS, significantly catering to the fine-grained interpretation needs of RS imagery in real-world applications.}}

\item{\textit{First RS Visual Prompt Learning Framework}. A universal region-level and point-level visual prompting data annotation method is developed. Subsequently, a sharing visual encoding mechanism is proposed to enhance the interplay among visual prompts, holistic images, and text instructions. Furthermore, the cross-domain learning strategy is designed, and the disjoint parameters are optimized in a lightweight manner by leveraging the multi-domain data,} endowing EarthMarker with spatial perception and joint instruction-following capabilities.

\item{\textit{First RS Visual Prompting Dataset, RSVP.} A large-scale RS regional instruction dataset named RSVP, containing roughly 3.65 M image-point-text and image-region-text pairings, is constructed for the first time. The construction of RSVP facilitates fine-grained RS imagery interpretation, laying the foundation for the development of visual prompting in the RS domain.}

\item{\textit{Superior Performance on Multi-granularity RS Visual Tasks}. Extensive experiments are conducted to demonstrate EarthMarker's competitive performance in multi-granularity RS visual interpretation tasks, compared with the SOTA specialist models, MLLMs, and visual prompting models. The tasks evaluated include scene classification, referring object classification, region captioning, inter-relationship analyses, etc. Therefore, EarthMarker successfully explores the adaptation of the visual prompt learning in the RS domain, improving the performance of MLLM and representing a significant step in fine-grained RS imagery interpretation.}
\end{itemize}

\section{Related Work}
\subsection{Multi-modal Large Language Models (MLLMs)}
Recently, the advancement of large language models (LLMs) has significantly fueled the revolution and innovation in the natural language processing (NLP) field. The representative works including closed-source GPT series~\cite{brown2020language,zhang2023gpt4roi} and open-source LLaMA series~\cite{touvron2023llama_a,touvron2023llama_b} have achieved powerful generalizable language processing and reasoning ability. Inspired by LLM and by further injecting visual signals, MLLMs are developed for visual-language mutual comprehension and various visual tasks. For example, VisualGPT~\cite{chen2022visualgpt}, BLIP~\cite{li2023blip} and Flamingo~\cite{alayrac2022flamingo} show strong multi-modal reasoning potential after aligning LLMs with visual modality. Notably, LLAMA-Adapter V2~\cite{gao2023llama} and SPHINX~\cite{lin2023sphinx} adopt zero-shot attention mechanism and linear projection layers tuning to mix LLM with visual signal. {Additionally, VisionLLM\cite{wang2024visionllm} aligns the definitions of vision-centric tasks with LLM methodologies to address vision-centric tasks in an open-ended manner.} Those nature scene MLLMs laid the foundation for the extension to the RS domain.

{Some pioneer RS MLLMs have emerged, and related studies such as EarthGPT~\cite{zhang2024earthgpt}, Geochat~\cite{kuckreja2024geochat}, SkyEyeGPT~\cite{zhan2024skyeyegpt}, and LHRS-Bot\cite{muhtar2024lhrs} have enabled MLLMs to interpret RS imagery. }Among them, Geochat is the first MLLM targeting solving multiple tasks on optical RS images. EarthGPT has proposed a universal MLLM that can deal with multi-source RS imagery and a wide range of RS visual tasks. {LHRS-Bot constructs the LHRS-Bench, to facilitate the RS community in evaluating the RS-specific MLLMs across diverse evaluation dimensions.} There is no doubt that those models facilitate the development of MLLMs in the RS-specific domain. {However, those models complete visual interpretation only through language interactions, but cannot generate responses guided by visual prompts to handle referring visual comprehension tasks. Apparently, existing RS MLLMs mainly focus on image-level and visual grounding, but are incapable of more fine-grained interpretation across region, point, and pixel levels. Therefore, this paper aims to enhance the MLLMs for the versatile multi-granularity understanding of RS imagery.}

\subsection{Prompt Engineering}
Prompt engineering is an emerging research direction in NLP~\cite{brown2020language}. Representation works contain AutoPrompt~\cite{shin2020autoprompt} and CoOp~ \cite{zhou2022learning}, which are designed to automate prompt template generation for language and vision-language models, instead of manual crafting. Additionally, language prompting has been applied for developing open-vocabulary detection models such as DetPro~\cite{du2022learning} and Promptdet~\cite{feng2022promptdet}. Compared with the extensively developed language prompting technique, visual prompting also needs more exploration. 

A major development is the SAM~\cite{kirillov2023segment} model, which supports multiple segmentation prompts to enhance the zero-shot performance. Due to the lack of semantic labels in SAM, the Semantic-SAM~\cite{li2023semantic} is proposed to realize multi-level semantics analysis and prediction. Notably, GPT4RoI~\cite{zhang2023gpt4roi} uses spatial boxes, and combines language and region-of-interest for input, enabling regional recognition. Colorful Prompting Tuning (CPT)~\cite{yao2022cptcolorfulprompttuning} uses color-based markers to improve the performance of pre-trained vision-language models. Note that Osprey~\cite{yuan2023osprey} incorporates masked regions into language instruction, achieving pixel-level visual understanding. Other visual prompting works including RegionBlip~\cite{zhou2023regionblip}, Kosmos-2~\cite{peng2023kosmos}, Shikra~\cite{chen2023shikra}, and Ferret~\cite{you2023ferret}, also have shown promising results in region-based image understanding by leveraging visual prompting techniques. The study entitled ``Visual Prompting via Image Inpainting"\cite{bar2022visual} shows that various vision tasks can be accomplished well by giving desired task examples. {Furthermore, RegionPLC\cite{yang2024regionplc} has proposed a lightweight and scalable regional point-language contrastive learning framework for open-world 3D scene understanding. However, the aforementioned models are trained on natural datasets, and the significant differences between natural and RS imagery suggest that their direct applicability to RS tasks is limited.} 

Limited studies in the RS domain have emerged on language or visual instruction learning recently, besides traditional vision models\cite{zhang2023posterior,10443257}. For example, RSVG~\cite{zhan2023rsvg} can provide the referred object’s locations based on natural language expression.
{EarthVQA\cite{wang2024earthvqa} dataset alongside a semantic object awareness framework (SOBA) is developed. EarthVQA contains images, corresponding semantic visual prompts, and QA pairs for training SOBA to achieve object-centric reasoning.}  Inspired by prompt learning, RSPrompter~\cite{chen2024rsprompter} designs an automated approach to generate appropriate prompts for SAM input, facilitating RS imagery segmentation. However, RSVG adopts language prompting but without visual prompting. EarthVQA solely involves pixel-level reasoning on city planning images, and RSPrompter is only tailored to prompt-based segmentation tasks. {Additionally, there are expert models that have developed detection and semantic segmentation with point supervision, including PSOD\cite{he2023learning,fang2023point,bearman2016s,mcever2020pcams}. Nevertheless, those models are designed for single tasks without versatile capabilities.} Obviously, there is no unified visual prompting framework designed for the RS domain to further improve the multi-granularity interpretation abilities of MLLMs. Those limitations hamper the development of more complex and fine-grained RS imagery understanding, therefore this paper focuses on filling this gap.

\begin{figure*}[!t]
	\centering
		\includegraphics[scale=0.14]{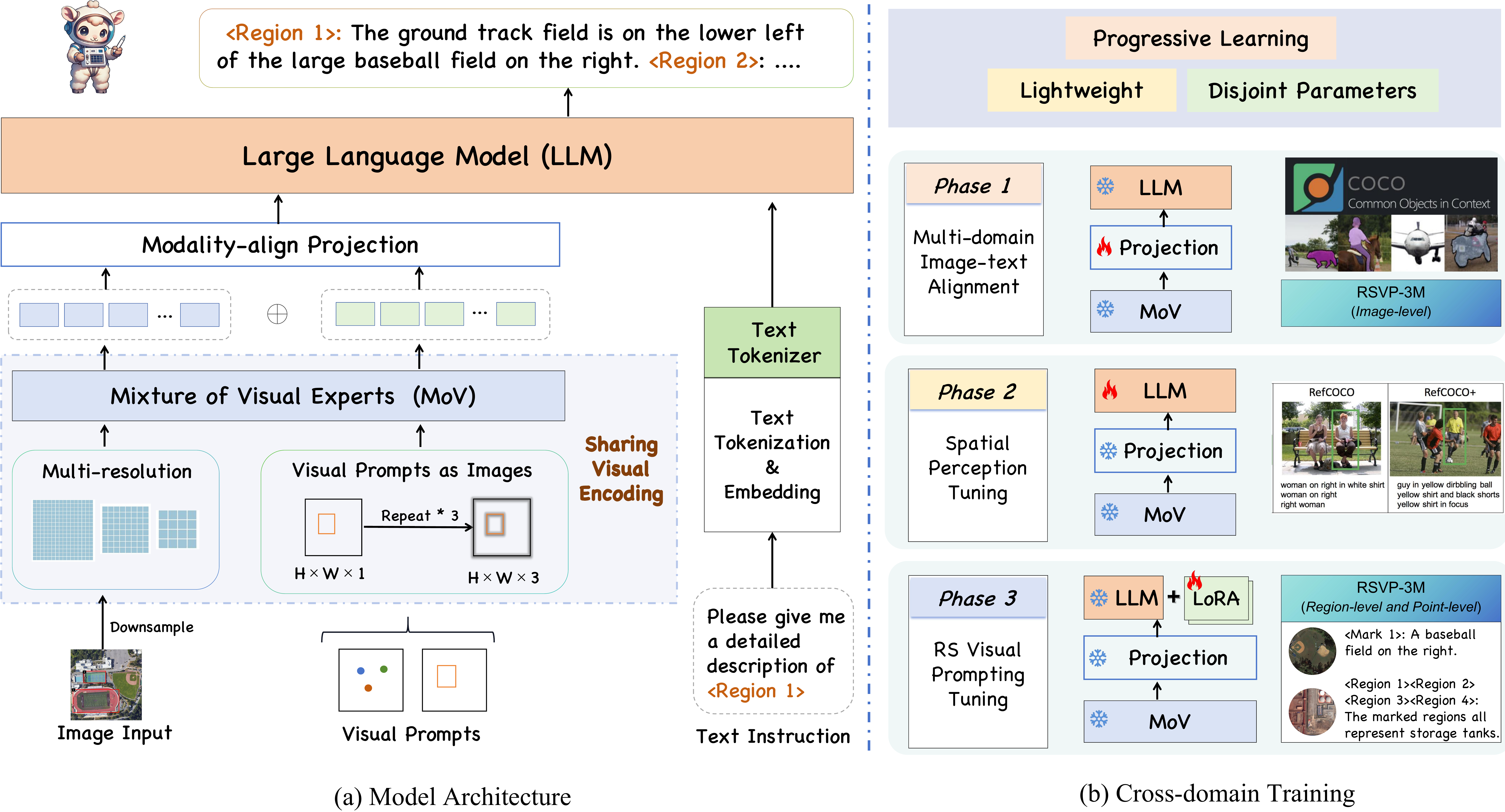}
	\caption{{(a) Overall model architecture of the proposed EarthMarker; (b) Cross-domain training.}}
    \label{FIG:model}
\end{figure*}

\section{Methodology}
We first overview the overall model architecture in Section III-A. Subsequently, the three-phase continuous training strategy of the proposed EarthMarker is detailed in Section III-B.

\subsection{Model Architecture}

In order to set up a flexible and effective interpretation way on complicated RS data, visual prompting-based MLLM is proposed. {As illustrated in Fig. \ref{FIG:model} (a), EarthMarker contains four core components: a sharing visual encoding module, a modality-align projection layer, a text tokenizer module, and an LLM decoder. }These components work together to deal with multi-modal information, such as text instruction, images, and diverse visual prompts including bounding boxes and points, allowing LLM to generate accurate text responses. Each part is introduced as follows in detail. 

{Theoretically, the first key point is establishing unified representation\cite{krishna2016visualgenomeconnectinglanguage,radford2021learning,komatsu2024a} across different modalities (e.g., text, images, and visual prompts), interpretation levels (e.g., image-level, region-level, and point-level), and a wide range of visual tasks. Specifically, all inputs $x_{\ddagger}$, where ${\ddagger} \in \{\mathrm{text},\mathrm{img},\mathrm{reg},\mathrm{pt}\}$, 
are embedded into the latent space, which is expressed as $\mathbf{z}_{\mathrm{text}} = f_{\mathrm{text}}(x_{\mathrm{text}}), ~\mathbf{z}_{\mathrm{img}} = f_{\mathrm{img}}(x_{\mathrm{img}}), ~\mathbf{z}_{\mathrm{reg}} = f_{\mathrm{reg}}(x_{\mathrm{reg}}),~\mathbf{z}_{\mathrm{pt}} = f_{\mathrm{pt}}(x_{\mathrm{pt}})
$, where $f_{\mathrm{text}}, f_{\mathrm{img}}, f_{\mathrm{reg}}, f_{\mathrm{pt}}$ are the respective embedding functions that map each input to a latent representation $\mathbf{z}_{\ddagger}$. Subsequently, the mathematics formula of the unified feature space processing can be expressed as
\begin{equation}
\mathcal{R} = \mathcal{G}_{\mathrm{uni}}(\mathbf{z}_{\mathrm{text}}, \mathbf{z}_{\mathrm{img}}, \mathbf{z}_{\mathrm{reg}}, \mathbf{z}_{\mathrm{pt}}),
\end{equation}
where $\mathcal{G}_{\mathrm{uni}}$ is a function that integrates text, image, region, and point-level information into a unified feature representation $\mathcal{R}$.}

{In EarthMarker, different modalities are processed through distinct encoder designs, ultimately achieving unified feature representation.} In particular, the images and corresponding visual prompts share a visual encoding mechanism for feature sharing, enabling the visual encoders to better understand and associate the relationship between images and visual prompts. Specifically, the Mixture of Visual Experts (MoV) \cite{gao2024sphinx} is designed to encode the visual information. The MoV incorporates two visual encoders, DINOv2-ViT L/14 \cite{oquab2023dinov2} and CLIP-ConvNeXt \cite{radford2021learning}, which are pre-trained on distinct network architectures, thus offering complementary visual semantics. To refine the robust multi-scale visual features, the input images $I$ are downsampled to different resolutions denoted as $I^i$ and then respectively fed into the MoV module to encode. Leveraging the strengths of various visual backbones, visual perception is enhanced and key details in images are refined. Subsequently, the encoded visual features are transformed to the same dimension and concatenated channel-wisely to obtain the integrated multi-scale feature maps represented as $\mathrm{V}_\text{img}$. This process can be formulated as 
\begin{equation}
\mathrm{V}_\mathrm{img} = \mathrm{Concat} ~(\mathrm{MoV}({I}^\mathrm{i})), ~\mathrm{i = 1,2,...,N}.
\end{equation}
{This approach refines the multi-scale visual information, enhancing the model’s perception across different resolutions.}

Notably, a key step to the encoder-sharing mechanism is called ``Visual Prompts as Images". Especially, the dimension ($\mathrm{H} \times \mathrm{W} \times 1$) of the visual prompts is processed to the same dimension ($\mathrm{H} \times \mathrm{W} \times 3$) with the images. {Then, the transformed visual prompts $\text{P}$ also can be fed into MoV together with the images, the encoded visual prompts expressed as $\mathrm{P'}_\text{j}$. Similarly, this process is written as
\begin{equation}
\mathrm{P'}_\text{j} = \mathrm{MoV} (\text{P}), ~\mathrm{j= 1, 2, 3},
\end{equation}
where $\mathrm{P'_{1}}$, $\mathrm{P'_{2}}$, and $\mathrm{P'_{3}}$ represents image-level, region-level, and point-level visual prompts, respectively.
}

Subsequently, the modality alignment projection layer $\Phi$ transforms the visual tokens into the language semantic space. Meanwhile, the text instructions are processed by the tokenizer module, which handles text tokenization and embedding, converting them into text embeddings $\mathrm{T_{instr}}$. After obtaining the projected image tokens, visual prompts tokes, and text instructions embeddings, they are integrated into an entire multi-modal input sequence. 

{To achieve cross-modal multi-granularity knowledge mutual enhancement \cite{you2023ferret}, the LLM decoder is designed as a bridge, linking various knowledge to ensure the model integrates a combination of global, regional, and detailed insights. LLM decoder takes the multi-modal inputs and dynamically adjusts the response, which features flexible task adaptation\cite{gao2024sphinx,wang2024multi}. Each task is triggered by text instructions $\mathrm{T_{instr}}$, which guide the model in adapting to different interpretation levels. The response process of the developed model is expressed through the following equations.}
{
\begin{equation}
\mathrm{L}_\mathrm{img}= \mathcal{M}_\mathrm{llm}(\Phi (\mathrm{V}_{\mathrm{img}}),\Phi (\mathrm{P'_{1}}),\mathrm{T_{instr}}),
\end{equation}
\begin{equation}
\mathrm{L}_\mathrm{reg}=\mathcal{M}_\mathrm{llm}(\Phi (\mathrm{V}_{\mathrm{img}}),\Phi (\mathrm{P'_{2}}),\mathrm{T_{instr}}),
\end{equation}
\begin{equation}
\mathrm{L}_\mathrm{pt}= \mathcal{M}_\mathrm{llm}(\Phi (\mathrm{V}_{\mathrm{img}}),\Phi (\mathrm{P'_{3}}), \mathrm{T_{instr}}),
\end{equation}
\begin{equation}
\mathrm{\mathcal{O}}_\mathrm{task} =\left\{\lambda_1\cdot\mathrm{L}_\mathrm{img}, ~\lambda_2\cdot\mathrm{L}_\mathrm{reg} ,~\lambda_3\cdot\mathrm{L}_\mathrm{pt}\right\},
\end{equation}
where $\lambda_1, \lambda_2, \lambda_3 \in \{0, 1\}$ indicate whether image-level, region-level, or point-level granularity is used for the task, $\mathrm{L}_\mathrm{img}, \mathrm{L}_\mathrm{reg}, \mathrm{L}_\mathrm{pt}$ represent the different levels interpretation results. $\mathcal{M}_\mathrm{llm}$ signifies the proposed model, $\mathrm{\mathcal{O}}_\mathrm{task}$ denotes the model output which dynamically adapts diverse tasks at different interpretation levels.}

\subsection{Cross-domain Training}

To bridge the domain gap between natural
and RS data, and effectively adapt general-domain knowledge into the RS domain, the cross-domain cross-domain learning training method is designed. {As depicted in Fig. \ref{FIG:model} (b), the entire training process is divided into three phases including multi-domain image-text alignment, spatial perception tuning, and RS visual prompting tuning stage.} Throughout the training, we keep lightweight training and avoid expensive full-parameters tuning. Furthermore, the disjoint parameters strategy is proposed, namely, the updated parameters of each stage are different. This strategy is conducive to the step-by-step solid understanding of images, and naturally solves the interference between image-text understanding, visual prompting comprehension, and multi-modal instruction-following.

\textbf{Multi-domain Image-text Alignment.} The first phase employs a multi-domain image-text alignment strategy. In this stage, both natural and RS domain image-level data are leveraged for pre-training to bring visual and text knowledge into alignment within a high-dimensional feature space. This strategy enables EarthMarker to deeply understand the holistic semantics of images. Specifically, we utilize the natural scene caption dataset COCO Caption~\cite{chen2015microsoft}, alongside the RS image caption and scene classification subset from the newly constructed RSVP. During this training phase, multi-scale visual features and language representations are integrated into the LLM to develop image-level comprehension capabilities. The MoV module is kept frozen throughout the training to concentrate on refining robust visual features. Only the alignment projection layer, which acts as the visual-language connector, undergoes parameter updates to enhance the multimodal capabilities of the proposed EarthMarker and ensure seamless integration of visual and textual information.

\textbf{Spatial Perception Tuning.} 
In the previous step, EarthMarker achieves image-level comprehension capability. In this step, to acquire spatial perception and object-level comprehension, the nature scene publicly available datasets, RefCOCO~\cite{kazemzadeh2014referitgame} and RefCOCO+ \cite{yu2016modelingcontextreferringexpressions} are utilized. Throughout the training, the attention layers of LLM are unfrozen for aligning spatial regional features with language embeddings. Specifically, LLM's key module self-attention head is composed of key $\mathrm{K}$, query $\mathrm{Q}$, and the value $\mathrm{V}$, which are transformed by several linear layers. The $l$-th implementation equations can be expressed as follows
\begin{equation}
\mathrm{Q}_l(X) = \mathrm{W}_l^{q} ~X +{\mathrm{b}_l^q},
\end{equation}
\begin{equation}
\mathrm{K}_l(X) = \mathrm{W}_l^{k} ~X +{\mathrm{b}_l^k},
\end{equation}
\begin{equation}
\mathrm{V}_l(X) = \mathrm{W}_l^{v} ~X +{\mathrm{b}_l^v},
\end{equation}
where $X$ represents multi-modal input. The parameters $\mathrm{W}_l^{q}$, $\mathrm{W}_l^{k}$, $\mathrm{W}_l^{v}$, $\mathrm{b}_l^q$, $\mathrm{b}_l^k$, and $\mathrm{b}_l^v$ are updated during the training. Then, the $l$-th single attention scores of $\mathrm{Q}$ and $\mathrm{K}$ are calculated as 
\begin{equation}
\text{Att}_l(\mathrm{Q} _l, \mathrm{K} _l, \mathrm{V} _l) = \mathrm{V} _l\times \text{Softmax}\left(\frac{\mathrm{Q} _l\mathrm{K} _l^T}{\sqrt{d_K}}\right),
\end{equation}
where $\sqrt{d_K}$ is the dimensionality of the keys. In addition, the other modules are kept frozen.

\textbf{RS Visual Prompting Tuning.} The last stage focuses on accurately following user visual instructions and achieving complex region-level and point-level visual reasoning tasks. The MoV, alignment projection, and LLM are fixed. The LoRA method is adopted for tuning. We load the weights trained in the previous phase and continue training EarthMarker on RSVP region-text and point-text parings, which contain the fine-grained referring object classification and referring brief caption data. Specifically, several learnable low-rank adapter matrices $\Delta{\mathrm{W} } _{h,l}^o,~ \Delta{\mathrm{W}_{h,l}^v}$, $\Delta{\mathrm{W}_{h,l}^q}$, and $\Delta{\mathrm{W}_{h,l}^k}$ are inserted into Transformer layers of LLM. The $H $ is the number of attention heads. The adapted multi-head attention is denoted as $\mathrm{MultiAttn}^*_l$, thus the output of the $l$-th adapted Transformer attention is formulated as 
\begin{align}
\mathrm{MultiAttn}^*_l =&\sum_{h=1}^{H}\left(\mathrm{O} _{h,l}+\Delta{\mathrm{W} } _{h,l}^o  \right)\times (\mathrm{V}_{h,l} \!+\! \Delta{\mathrm{W}_{h,l}^v})\\
&\times\! \text{Softmax}\!\!\left(\!\frac{(\mathrm{Q}_{h,l} + \Delta{\mathrm{W}_{h,l}^q})(\mathrm{K}_{h,l}^T \!+ \!\Delta{\mathrm{W}_{h,l}^k})}{\sqrt{d_K}}\!\right)\!\!  \nonumber.
\end{align}

In conclusion, the cross-domain training endows EarthMarker with various granular (e.g., image-level, point-level, and region-level) multi-modal instruction capabilities in the RS domain. In the first multi-domain image-text alignment stage, the LLM is efficiently converted into an MLLM, which is capable of image-level understanding. Subsequently, by utilizing the nature scene referring datasets, EarthMarker is equipped with the fundamental spatial perception of images. This is beneficial for subsequent developments of referring comprehension ability in the RS domain. {Most importantly, the training on large-scale natural scenes lays the foundation for EarthMarker to achieve the adaptation to diverse resolutions imagery, and real-world scenarios. Furthermore, by leveraging the RS visual prompting dataset RSVP, EarthMarker is endowed with image understanding across the image, region, and point levels.} Notably, different field datasets are adopted for training, and enhancing open-vocabulary reasoning ability. It should be emphasized that during the whole training, our updatable parameters are disjoint, preventing interference between understanding images at different granularity and the capability to follow visual prompts. {After step-wise progressive learning, EarthMarker is capable of flexibly shifting focus between global context and fine details as needed.}

\section{RS Visual Prompting Dataset Construction}

\begin{table}[!b]
\centering
\renewcommand{\arraystretch}{1.7}
\caption{{Details on the training samples of RSVP.}}
\label{tab:MMRS-1M Components}
\setlength{\tabcolsep}{2pt}
\scalebox{1}{
\fontsize{8pt}{8pt}\selectfont

\begin{tabular}{cccc}
\hline
{Tasks}                                    & Raw Datas                & Samples   & Resolutions \\ \hline
\multirow{3}{*}{\begin{tabular}[c]{@{}c@{}}{\shortstack{Image Captioning}}\end{tabular}}          
                                        & NWPU-Captions~\cite{cheng2022nwpu}              & 169,981 & 0.2$-$30m \\ \cline{2-4} 
                                        & RSITMD\cite{yuan2022exploring}                     & 24,387 & $-$ \\ \cline{2-4} 
                                        & Sydney-Captions~\cite{qu2016deep}            & 2,837 & 0.5m \\ \cline{2-4} 
                                           \hline
\shortstack{Region Captioning}                 & DIOR-RSVG~\cite{10056343}                  & 31,491 & 0.5$-$30m  \\ \hline
\multirow{4}{*}{\shortstack{Scene Classification}}         & NWPU-RESISC45~\cite{7891544}              & 94,500 & 0.2$-$30m \\ \cline{2-4} 
                                        & OPTIMAL 31~\cite{wang2018scene}                    & 3,720 & $-$ \\ \cline{2-4} 
                                        & RSD46~\cite{long2017accurate}           & 350,685 & 0.5$-$2m \\ \cline{2-4} 
                                        & WHU-RS19~\cite{dai2010satellite}                   & 3,015 & $< 0.5$m \\ \cline{2-4} 
                                        \hline
                                        
\multirow{11}{*}{\shortstack{Referring Object \\ Classification (Box)}}             
                                        & DOTA V2~\cite{9560031}                       & 99,774 & $-$ \\ \cline{2-4} 
                                        & FAR1M~\cite{sun2022fair1m}                      & 17,074 & 0.3$-$0.8m \\ \cline{2-4} 
                                        & NWPUVHR10~\cite{cheng2016learning}                  & 1,888 & 0.5$-$2m \\ \cline{2-4}  
                                        & RSOD~\cite{long2017accurate}                       & 1,465 & 0.3$-$3m \\ \cline{2-4} 
                                        & UCAS-AOD~\cite{zhu2015orientation}                   & 3,203 & $-$ \\ \cline{2-4} 
                                        & VisDrone~\cite{zhu2021detection}                   & 128,531 & $-$ \\ \cline{2-4}  
                                        & MAR20~\cite{wenqi2024mar20}                       & 2,096 & $-$ \\ \cline{2-4} 
                                        & DOSR~\cite{han2021fine}                          & 1,127 & 0.5$-$2.5m \\ \cline{2-4} 
                                        & LEVIR~\cite{zou2017random}                    & 4,633 & 0.2$-$1m \\ \cline{2-4} 
                                        & HRSC2016~\cite{liu2017high}               & 2,171 & 0.4$-$2m \\ \cline{2-4} 
                                        & HRRSD~\cite{zhang2019hierarchical}          & 12,519 & 0.15$-$1.2m \\ \hline
                                        
\multirow{9}{*}{\shortstack{Referring Object \\ Classification (Point)}}            
                                        & Vaihingen~\cite{ISPRS}                    & 45,104 & 0.09m \\ \cline{2-4}  
                                        & Potsdam~\cite{ISPRS}                     & 504,139 & 0.05m \\ \cline{2-4}   
                                        & Hi-UCD~\cite{tian2020hi}                 & 125,908 & 0.1m \\ \cline{2-4}   
                                        & NWPUVHR10~\cite{cheng2016learning}                 & 956 & 0.5$-$2m \\ \cline{2-4}   
                                        & isAID~\cite{waqas2019isAID}                & 294,355 & $-$ \\ \cline{2-4}   
                                        & UAVid~\cite{LYU2020108}               & 1,167,543 & $-$ \\ \cline{2-4}   
                                        & SOTA~\cite{wang2024samrs}              & 171,583 & $-$ \\ \cline{2-4}   
                                        & FAST~\cite{wang2024samrs}                  & 263,056 & 0.3$-$0.8m \\ \cline{2-4}   
                                        & WHU~\cite{ji2018fully}                & 33,022 & 0.3m \\ \hline
\multirow{6}{*}{\shortstack{Relationship  Analyses}}            
                                        & HRRSD~\cite{zhang2019hierarchical}                 &  11602  & 0.15$-$1.2m \\ \cline{2-4}  
                                        & RSOD~\cite{long2017accurate}                   & 934 & 0.3$-$3m  \\ \cline{2-4}  
                                        & NWPUVHR10~\cite{cheng2016learning}                   & 1,950 & 0.5$-$2m \\ \cline{2-4}  
                                        & LEVIR~\cite{zou2017random}                 & 3,956 &  0.2$-$1m\\ \cline{2-4}  
                                        & UCAS-AOD~\cite{zhu2015orientation}                  & 2,892 & $-$ \\ \cline{2-4}  
                                        & DIOR-RSVG~\cite{10056343}                 & 15,274 & 0.5$-$30m \\ \hline
\multirow{3}{*}{\shortstack{Grounded  Captioning}}  
                                        & DIOR-RSVG~\cite{10056343}                    & 15,274 & 0.5$-$30m \\ \cline{2-4}
                                        & GeoChat~\cite{kuckreja2024geochat}                  & 35,264 & $-$ \\ \cline{2-4}
                                        & NWPUVHR10~\cite{cheng2016learning}                    & 975 & 0.5$-$2m \\ \hline
\end{tabular}
\label{datasets}
}
\end{table}

In this section, a visual prompting dataset called RSVP is presented. RSVP is the first visual prompting instruction dataset in the RS field. {EarthMarker is designed to advance coarse-grained image-level and fine-grained point/region-level RS-specific MLLMs.} Specifically, RSVP contains roughly 3.65 million multi-modal dialogue data with visual prompts. {Those multi-granularity visual prompting data are restructured and cleaned from existing publicly available RS datasets, covering global data distribution and different spatial resolutions.} Furthermore, the GPT4V\cite{achiam2023gpt} is employed for automatic annotation to construct a high-quality complex visual reasoning dataset\cite{lin2024draw}. A detailed explanation of the construction of the RSVP dataset is introduced as follows.

\subsection{Data Conversion and Annotation from Public RS Datasets}

A part of the data of the dataset RSVP is constructed by restructuring and relabeling existing RS datasets. {As shown in Tab. \ref{tab:MMRS-1M Components}, a range of visual task types are covered, containing scene classification, referring object classification, image captioning, region captioning, relationship analyses, etc. Notably, RSVP includes various RS images that feature different spatial resolutions, containing very high-resolution (VHR) optical images and medium- to low-resolution imagery. In addition, most of the raw data comes from Google Earth which is sourced from various satellites or aerial photographs, providing broad geographic coverage.} 

In RSVP, each data item consists of visual prompts, user instructions, and an image. The visual prompts in the user instructions or model answers are expressed as $\mathrm{<Mark~i>}$ or $\mathrm{<Region~i>}$. {Notably, when performing image-level tasks, the $\mathrm{<Region~i>}$ covers the entire image.} The image-level, region-level, and point-level data are derived from different RS dataset types. For example, image-level visual prompting data is converted from image classification and captioning datasets. {For the two type datasets, the bounding box $[0, 0,\mathrm{width},\mathrm{height}]$ served as the image-level visual prompt, which represents the entire image. For image-level detailed captioning tasks, the text instruction is ``Please provide a detailed description of the $\mathrm{<Region~i>}$ in the image" to guide the model for image-level captioning.}

{The region-level data is based on object detection datasets. The ground truth bounding boxes are used as visual prompts to instruct the model to identify the object-level or region-level categories accurately. Take the scene of the airport as an example, the text instruction for the region captioning task is ``{Please provide the brief caption of each marked region in the image}, and the corresponding answer format generated by the model is ``$<\rm\! Region~1\!>: \mbox{A big airplane on the left} \backslash{\rm n} < {\rm \!Region~2\!>:}$ ${\rm{\mbox{A small vehicle on the top}}\backslash{\rm n},...,}${`bbox}’$:[x_1,y_1,x_2,y_2],...$".}

Notably, the point-level data is transformed from segmentation datasets. For instance segmentation, the representative points extracted from masks corresponding to instances are used as point-level visual prompts. For semantic segmentation, each image is divided into $32 \times 32$ patches, and the points are randomly sampled within each patch as the visual prompts, with the category retrieved from the corresponding segmentation map. {The user instruction which guides the model to accomplish point-level referring object classification task is ``{Please identify the category of each marked point in the image}". The answer format of point-level is ``$<{\rm\! Mark~1\!>: Label~1}\backslash{\rm n} < {\rm \!Mark~2\!>: Label~2}\backslash{\rm n},...,${`points}’$:[x_1,y_1],[x_2,y_2],...$". }

The data structures of other visual tasks are similar to those explained above. Through the transformation and re-annotation based on public datasets, the visual prompting dataset RSVP is effectively developed, containing image-point-text and image-region-text pairings.

\subsection{GPT4V-assisted Visual Prompting Data Generation}
The aforementioned public datasets only provide simple classification information and brief captions, which are insufficient for intelligent interpretation of complex RS imagery. To mitigate the limitation and develop a more detailed and explicit RS visual prompting dataset, the language prompts for GPT4V are carefully crafted for generating data featuring various complex visual reasoning. {The complex fine-grained visual tasks involve detailed region captioning, object inter-relationship analysis, grounding captioning, etc.} We adopt the Set-of-Marks\cite{yang2023set} (SoM) prompting, which can effectively unleash the extraordinary visual grounding ability of GPT4V, to obtain comprehensive and unique characteristics from the RS imagery.

The data generated using GPT4V not only compensates for the lack of information in brief captions but also provides detailed descriptions that reveal the spatial and semantic relationships between different regions in the image. For example, in aerial imagery, it is feasible to identify the general category of the image and provide a simple description. Additionally, detailed descriptions, such as the spatial layout of tennis courts, basketball courts, and playgrounds, the relationships among these areas, and the activities of people on the playground, can be conducted. The RSVP dataset, supplemented by public datasets and data generated by GPT4V, covers a wide range of fine-grained visual reasoning tasks, enhancing the richness and diversity of the data. 
\vspace{0.5cm}
\section{Experiments}
In this section, we present extensive experiments to validate the superior performance of EarthMarker. {In Section V-A, we introduce the implementation setups including evaluation metrics and implementation details.} Subsequently, we conduct qualitative analyses to provide a holistic view of EarthMarker's performance from Sections V-B to V-E. 
 
{\subsection{Implementation Setups}
\subsubsection{Evaluation Metrics}
For scene classification tasks, we utilize the scene classification accuracy metric, which specifically measures the proportion of correctly classified scenes out of the total number of scenes. Subsequently, for the evaluation of image and region captioning tasks, we employ a range of metrics designed to assess the quality of language generation. These metrics include BLEU scores\cite{papineni2002bleu}, which is a modified precision metric with a sentence-brevity penalty, calculated as a weighted geometric mean over different length n-grams. METEOR \cite{banerjee2005meteor} uses exact, stem, synonym, and paraphrase matches between n-grams to align sentences, before computing a weighted F-score with an alignment fragmentation penalty; ROUGE-L \cite{lin2004rouge} focuses on the longest common subsequence to assess the fluency and relevance of the captions; CIDEr\cite{vedantam2015cider}, applies term frequency-inverse document frequency (TFIDF) weights to n-grams in the candidate and reference sentences, which are then compared by summing their cosine similarity across n-grams; and SPICE\cite{anderson2016spice}, which assesses the quality of image captions, by comparing the generated captions with reference captions the based on semantic similarity. Furthermore, for referring object classification tasks evaluation, following\cite{yuan2023osprey,lin2024draw}, we use Semantic Intersection over Union (S-IOU) and Semantic Similarity (SS)\cite{rezatofighi2019generalized}. SS measures the similarity of predicted and ground truth labels in a semantic space, while S-IOU reflects the overlap of words.}

\subsubsection{Implementation Details}
The proposed EarthMarker adopts the cross-domain training strategy, and the parameters updated vary at different stages. In general, we train an off-the-shelf 13B language model Llama 2 and the visual encoder MoV is kept frozen during the training. In the first multi-domain image-text alignment stage, only the alignment projection layer is updated. Then, in the spatial perception tuning phase, only the attention layers of LLM are unfrozen. Furthermore, the trainable LoRA metrics are introduced in the last RS visual prompting tuning stage. {We utilize AdamW optimizer\cite{kingma2014adam} with weight decay = 0 and betas = (0.9, 0.95), the learning rate is set to 2e-5, and the total training stages are conducted on 8 NVIDIA A100 GPUs (each equipped with 80 GB of memory) and the training time is around 316 hours.}

For model evaluation, we select diverse multi-granularity visual tasks to assess the performance of EarthMarker. Image-level tasks include scene classification and image captioning, while region-level tasks contain referring object classification, region captioning, and relationship analyses. {In addition, we evaluate EarthMarker under both zero-shot and supervised settings to compare its performance with other existing models. To fully validate the superiority of EarthMarker, we select a range of models, including natural scene MLLMs, RS-specific MLLMs, and expert models. Note that the newly constructed RSVP contains only training samples for developing visual prompting MLLMs and does not include a test set. For the test sets, we use new datasets not involved in RSVP or test subsets from raw datasets, depending on the specific task, to ensure a comprehensive evaluation. Specifically, in the scene classification task, we use a zero-shot setting with new datasets such as AID \cite{xia2017aid} and UCMerced \cite{yang2010bag}. For other tasks, we use a supervised setting and evaluate the model performance on the test sets of DIOR-RSVG, etc. More descriptions of model selection and settings are detailed in Sections V-B to V-E.}

\begin{table}[!t]
\caption{{Zero-shot comparison results between EarthMarker and other MLLMs on UCMerced and AID. }}\label{tab:Classification_compare_zeroshot}
\centering

\renewcommand{\arraystretch}{1.3}
\scalebox{1}{
\fontsize{10pt}{10pt}\selectfont
\begin{tabular}{lcc}  
\toprule      
\multicolumn{1}{c}{Models}          & UCMerced & AID   \\ 
\cmidrule(lr){1-3}
\multicolumn{1}{l}{Qwen-VL-Chat~\cite{bai2023qwen}} & 62.90& 52.60             \\
\multicolumn{1}{l}{MiniGPTV2~\cite{chen2023minigpt}}    & 4.76    & 12.90                \\
\multicolumn{1}{l}{LLaVa-1.5~\cite{liu2023visual}}    & 68.00    & 51.00               \\
\multicolumn{1}{l}{Sphinx~\cite{lin2023sphinx}}      & 62.76   & 58.20               \\
\multicolumn{1}{l}{GeoChat~\cite{kuckreja2024geochat}}      & 84.43   & 72.03              \\
\cmidrule(lr){1-3}
\multicolumn{1}{l}{\textbf{EarthMarker (Ours)}} & \textbf{86.52}    & \textbf{77.97}        \\ 
\bottomrule
\end{tabular}
} 
\end{table}

\begin{table*}[!b]
\caption{{Supervised comparison results on NWPU-captions dataset between expert models and our EarthMarker. }}\label{tab:NWPU_caption_compare_supervised}
\centering

\renewcommand{\arraystretch}{1}
\setlength{\tabcolsep}{3pt}
\scalebox{1}{
\fontsize{10pt}{10pt}\selectfont
\begin{tabular}{lccccccccc}
\toprule
\multicolumn{1}{c}{Models} & BLEU1 & BLEU2 & BLEU3 & BLEU4 & METEOR & ROUGE-L & CIDEr(0-5) & SPICE \\ 
\cmidrule(lr){1-9}
\textit{\textbf{Expert Models}}        &       &       &       &       &        &         &            &       \\ 
\multicolumn{1}{l}{CSMLF~\cite{8633358}}                 & 77.0 & 64.9 & 53.2 & 47.1 & 32.0  & 57.8   & 106.5      & 26.5 \\
\multicolumn{1}{l}{Qu, \textit{et. al}~\cite{qu2016deep}}           & 72.5 & 60.3 & 51.8 & 45.5 & 33.6  & 59.1   & 117.9      & 27.6 \\
\multicolumn{1}{l}{Attention (soft)~\cite{lu2017exploring}}      & 73.1 & 60.9 & 52.5 & 46.2 & 33.9  & 59.9   & 113.6      & \textbf{28.5} \\
\multicolumn{1}{l}{Attention (hard)~\cite{lu2017exploring}}       & 73.3 & 61.0 & 52.7 & 46.4 & 34.0  & 60.0   & 110.3      & 28.4 \\
\multicolumn{1}{l}{FC-Att~\cite{zhang2019description}}           & 73.6 & 61.5 & 53.2 & 46.9 & 33.8  & 60.0   & 123.1      & 28.3 \\
\multicolumn{1}{l}{SM-Att~\cite{zhang2019description}}           & 73.9 & 61.7 & 53.2 & 46.8 & 33.0  & 59.3   & 123.6      & 27.6 \\
\multicolumn{1}{l}{MLCA-Net~\cite{9866055}}              & 74.5 & 62.4 & 54.1 & 47.8 & 33.7  & 60.1   & 126.4      & \textbf{28.5} \\ 
\cmidrule(lr){1-9}
\textit{\textbf{Visual Prompting Model}}                 \\ 
\multicolumn{1}{l} {\textbf{EarthMarker (Ours)}} & \textbf{84.4} & \textbf{73.1} & \textbf{62.9} & \textbf{54.3} & \textbf{37.5} & \textbf{70.0} & \textbf{162.9} & 26.8 \\

\bottomrule
\end{tabular}
}
\end{table*}
\subsection{Scene Classification}

For scene classification tasks\cite{zhuang2024heterogeneous}, we use the AID\cite{xia2017aid} and UCMerced\cite{yang2010bag} datasets for evaluation. AID is a large-scale aerial dataset collected from Google Earth, containing 30 categories. {Following the setting of GeoChat\cite{kuckreja2024geochat}, we use a 20$\%$ split of the AID dataset for testing. Moreover, the entire UCMerced dataset is adopted as a zero-shot test set, which consists of 21 categories for scene classification.} 

We prompt the model with an image-level box $[0, 0,\mathrm{width},\mathrm{height}]$ to represent the entire image. The text instruction is ``Please identify the object category of each marked region in the image". We calculate the zero-shot accuracy on the AID and UCMerced dataset. EarthMarker significantly outperforms other MLLMs, with an accuracy of 86.52$\%$ on UCMerced and 77.97$\%$ on AID, as presented in Tab. \ref{tab:Classification_compare_zeroshot}. In comparison, LLaVa-1.5 and Sphinx, due to the lack of RS domain knowledge, are inferior to the RS MLLM GeoChat and our EarthMarker. Compared to GeoChat, our EarthMarker achieves an accuracy improvement of 5.94$\%$ on AID and 2.09$\%$ on UCMerced. Owing to the multi-domain image-text alignment training, EarthMarker is endowed with excellent holistic scene understanding ability on RS imagery.

\begin{table}[!t]
\caption{{Comparison results between EarthMarker and other MLLMs, visual prompting models on the test set of DIOR-RSVG. }}\label{tab:Referring_classification_compare}

\centering
\renewcommand{\arraystretch}{1.3}
\scalebox{1}{
\fontsize{10pt}{10pt}\selectfont
\begin{tabular}{lccc}  
\toprule      
\multicolumn{1}{c}{Models}          & \multicolumn{1}{l|}{Formats} & SS & SIOU   \\ 
\cmidrule(lr){1-2}\cmidrule(lr){3-4}
\textit{\textbf{MLLMs}}        &       &       &      \\              
\multicolumn{1}{l}{GeoChat~\cite{kuckreja2024geochat}} & \multicolumn{1}{l|}{Coor}       & 79.59& 68.80             \\
\multicolumn{1}{l}{Sphinx~\cite{lin2023sphinx}}    & \multicolumn{1}{l|}{Coor}       & 93.72    & 89.37                \\
\multicolumn{1}{l}{EarthGPT~\cite{zhang2024earthgpt}}    & \multicolumn{1}{l|}{Coor}      & 94.64   & 90.16              \\ 
\cmidrule(lr){1-2}\cmidrule(lr){3-4}
\textit{\textbf{Visual Prompting Models}}        &       &       &      \\    
\multicolumn{1}{l}{Sphinx-V~\cite{lin2024draw}}      & \multicolumn{1}{l|}{Box}        & 89.07   & 81.62             \\
\multicolumn{1}{l}{Vip-LLava-7b~\cite{cai2024vip}}      & \multicolumn{1}{l|}{Box}         & 72.56   & 55.94            \\
\multicolumn{1}{l}{Vip-LLava-13b~\cite{cai2024vip}}      & \multicolumn{1}{l|}{Box}     & 74.51   & 60.53           \\
\cmidrule(lr){1-2}\cmidrule(lr){3-4}
\multicolumn{1}{l}{\textbf{EarthMarker (Ours)}} &\multicolumn{1}{l|}{Point}             & \textbf{95.96}    & \textbf{93.49}        \\ 
\multicolumn{1}{l}{\textbf{EarthMarker (Ours)}} &\multicolumn{1}{l|}{Box}            & \textbf{98.37}    & \textbf{97.24}        \\ 
\bottomrule
\end{tabular}
} 
\end{table}

\begin{table*}[!t]
\caption{{Comparison results between EarthMarker and other MLLMs, visual prompting models on the test set of DIOR-RSVG. }}\label{tab:Region_caption_compare_dior}
\centering

\renewcommand{\arraystretch}{1.3}
\scalebox{0.95}{
\fontsize{10pt}{10pt}\selectfont
\begin{tabular}{lccccccccc}  
\toprule      
\multicolumn{1}{c}{Models}          & \multicolumn{1}{l|}{Formats} & BLEU-1 & BLEU-2& BLEU-3& BLEU-4& METEOR & ROUGE &CIDEr &SPICE\\ 
\cmidrule(lr){1-2}\cmidrule(lr){3-10}
\textit{\textbf{MLLMs}}        &       &       &   &    &       &       & &       &    \\      
\multicolumn{1}{l}{Qwen-VL-Chat~\cite{kuckreja2024geochat}} & \multicolumn{1}{l|}{Coor}       & 20.66	&9.55	&4.90&	1.96&	8.35&	20.93&	29.18	&9.55             \\
\multicolumn{1}{l}{GeoChat~\cite{bai2023qwen}} & \multicolumn{1}{l|}{Coor}       & 20.86&	9.63&	5.43&	3.25&	12.94&	26.98&	30.92&	24.97           \\
\multicolumn{1}{l}{Sphinx~\cite{lin2023sphinx}}    & \multicolumn{1}{l|}{Coor}       & 43.32	&33.58&	27.58&	22.81	&21.70	&47.81	&235.07	&38.12                \\
\multicolumn{1}{l}{EarthGPT~\cite{zhang2024earthgpt}}    & \multicolumn{1}{l|}{Coor}      & 49.73&	39.24&	32.50&	26.79&	24.09&	47.87&	232.79	&38.29             \\ 
\cmidrule(lr){1-2}\cmidrule(lr){3-10}
\textit{\textbf{Visual Prompting Models}}        &       &   &    &       &       & &       &    \\  
\multicolumn{1}{l}{Sphinx-V~\cite{lin2024draw}}      & \multicolumn{1}{l|}{Box}        & 45.19	&34.83	&28.59&	23.64&	24.34&	50.54&	235.09&	43.53             \\
\multicolumn{1}{l}{Vip-LLava-7b~\cite{cai2024vip}}      & \multicolumn{1}{l|}{Box}         & 21.03&	11.26	&6.37&	3.25&	9.59&	23.50&	30.95&	12.44            \\
\multicolumn{1}{l}{Vip-LLava-13b~\cite{cai2024vip}}      & \multicolumn{1}{l|}{Box}     & 21.68	&11.17&	6.23	&3.38&
	9.18	&23.22&	28.58	&10.68           \\
\multicolumn{1}{l}{GLaMM~\cite{cai2024vip}}      & \multicolumn{1}{l|}{Box}     & 23.30	&14.18&	8.78	&5.15&	11.24	&30.36	&64.08	&17.99           \\
\cmidrule(lr){1-2}\cmidrule(lr){3-10}
 \multicolumn{1}{l}{\textbf{EarthMarker(Ours)}} & \multicolumn{1}{l|}{Point}            & \textbf{57.45}	& \textbf{49.23}	& \textbf{43.88}	& \textbf{39.46}	& \textbf{32.26}	& \textbf{61.51}	& \textbf{400.76}	& \textbf{61.03}       \\ 
 \multicolumn{1}{l}{\textbf{EarthMarker(Ours)}} & \multicolumn{1}{l|}{Box}            & \textbf{57.14}	& \textbf{48.60}	& \textbf{43.06}	& \textbf{38.59}	& \textbf{31.97}	& \textbf{60.46}	& \textbf{379.25}	& \textbf{59.87}        \\ 
\bottomrule
\end{tabular}
} 
\end{table*}

\begin{table*}[!h]
\centering 
\caption{{Ablation results about the design of the visual encoders. }}\label{tab:Ablation on visual encoder}
\renewcommand{\arraystretch}{1.3}
\setlength{\tabcolsep}{2pt}
\scalebox{0.95}{
\fontsize{10pt}{10pt}\selectfont

\begin{tabular}{ccccccc}
\toprule  
\multicolumn{2}{c}{CNN Model}  & \multicolumn{1}{c|}{ViT Model}       & Classification             & Region Captioning            & Referring Object Classification  \\ 
\cmidrule(lr){1-2}\cmidrule(lr){3-3}\cmidrule(lr){4-6}
CLIP ConvNeXt-L  &  CLIP RN50$\times$16 & \multicolumn{1}{c|}{DINOv2 ViT-L/14} & AID (Top-1 Acc) & DIOR-RSVG (CIDEr) & DIOR-RSVG (SIOU)  \\ 
\cmidrule(lr){1-2}\cmidrule(lr){3-3}\cmidrule(lr){4-6}
\checkmark             &   $\boldsymbol\times $            &      \multicolumn{1}{c|}{$\boldsymbol\times $ }             & 73.83                     & 368.21           & 94.27         \\
$\boldsymbol\times $                 & \checkmark          &      \multicolumn{1}{c|}{$\boldsymbol\times $ }              & 74.09                    & 368.10           & 94.52         \\
$\boldsymbol\times $                 & $\boldsymbol\times $              & \multicolumn{1}{c|}{\checkmark}              & 73.21               & 369.84                 & 95.93         \\
 $\boldsymbol\times $                & \checkmark          & \multicolumn{1}{c|}{\checkmark}              & 75.39               & 372.41                 & 96.81        \\
\checkmark             &  $\boldsymbol\times $             & \multicolumn{1}{c|}{\checkmark}              & \textbf{77.97}                     & \textbf{379.25}           & \textbf{97.24}         \\ 
\bottomrule
\end{tabular}
}
\end{table*}

\begin{table*}[!h]
\centering  
\caption{{Ablation results about the composition of training data. }}\label{tab:Ablation on training data}
\renewcommand{\arraystretch}{1.4}
\setlength{\tabcolsep}{2pt}
\scalebox{0.84}{
\fontsize{10pt}{10pt}\selectfont

\begin{tabular}{cccc|cccccc}
\toprule  
 \multicolumn{4}{c|}{Training Data Composition}       & {Classification}             & Region Captioning            & Referring Object Classification  \\ 
\cmidrule(lr){1-4}\cmidrule(lr){5-7}\cmidrule(lr){8-10}
Natural Caption & Natural Referring  & RS Reconstructed & \multicolumn{1}{c|}{GPT4V Generated} & AID (Top-1 Acc) & DIOR-RSVG (CIDEr) & DIOR-RSVG (SIOU)  \\ 
\cmidrule(lr){1-1}\cmidrule(lr){2-2}\cmidrule(lr){3-4}\cmidrule(lr){5-7}
$\boldsymbol\times$ & $\boldsymbol\times $ & \checkmark & $\boldsymbol\times$ & 75.24 & 352.82 & 96.46 \\
\checkmark & $\boldsymbol\times$ & \checkmark & $\boldsymbol\times$ & 76.97 & 356.25 & 97.01 \\ 
\checkmark & \checkmark& \checkmark  & $\boldsymbol\times $ & 77.94 & 358.50 & 97.20 \\
\checkmark  & \checkmark & \checkmark & \checkmark  & \textbf{77.97} & \textbf{379.25} & ~\textbf{97.24 } \\ 
\bottomrule
\end{tabular}
}
\end{table*}

\subsection{Image Captioning}
To evaluate the image captioning capabilities, we use the NWPU-Captions\cite{cheng2022nwpu} dataset to assess and compare EarthMarker against other expert models in the supervised setting. Created by Northwestern Polytechnical University, the NWPU-Captions dataset includes 31,500 aerial images and 157,500 sentences for RS image description. {Following the protocol of MLCA-Net~\cite{9866055,zhang2024earthgpt}, we employ BLEU1, BLEU2, BLEU3, BLEU4, METEOR, ROUGE-L, and CIDErD as evaluation metrics, which offers a comprehensive language quality evaluation.} In the evaluation, we use the sentence ``Please provide a brief caption of each marked region in the image." as text instruction and a full-image box $[0, 0,\mathrm{width},\mathrm{height}]$ as the visual prompt.  As shown in Tab. \ref{tab:NWPU_caption_compare_supervised}, compared to other expert models, EarthMarker demonstrates improvements in BLEU1, BLEU2, BLEU3, BLEU4, METEOR, and ROUGE-L by 7.4$\%$, 8.2$\%$, 8.8$\%$, 6.5$\%$, 3.5$\%$, and 9.9$\%$,  respectively, and a 36.5$\%$ improvement in CIDErD. In summary, we dexterously annotate the entire image as a single region, and further using both natural and RS image training, EarthMarker achieves a high level of understanding of the overall image, outperforming expert models.

\subsection{Referring Object Classification}

The referring object classification task aims to identify the category within the referring region in the image. {Following \cite{lin2024draw,yuan2023osprey}, the metrics used to evaluate this task are two semantic relevance indicators—Semantic Similarity (SS) and Semantic Intersection over Union (S-IOU) to assess a model’s object classification ability. }Closed-set testing is adopted on the test sets of object-level dataset DIOR-RSVG~\cite{10056343}. The text instruction is ``Please identify the category of the marked region in the image", which along with the bounding boxes are fed into LLM to predict the category of regions. Due to the previous MLLM (e.g., GeoChat, Sphinx, and EarthGPT) only accepting images and text as input, the region prompts for those MLLMs are coordinates information contained in the text instructions. 

As the results shown in Tab. \ref{tab:Referring_classification_compare}, EarthMarker achieves 95.96 $\%$ in SS and 93.49 $\%$ in S-IoU using point-level visual prompts, and 98.37$\%$ in SS and 97.24$\%$ in S-IoU based on box-level visual prompts on the DIOR-RSVG dataset. Both sets of results significantly outperform the SOTA method. Furthermore, EarthMarker surpasses the previous SOTA model EarthGPT by 3.73$\%$ in SS and 7.08$\%$ in S-IoU on the DIOR-RSVG dataset, demonstrating its superior capability in fine-grained box-level classification and the effectiveness of our designed visual prompting method. 

\subsection{Region Captioning}

For the brief region captioning, the test set of DIOR-RSVG~\cite{10056343} is employed. Specifically, we adopt boxes as the visual prompt and a text prompt, such as ``Please provide a brief caption of each marked region in the image," to prompt EarthMarker to concisely describe the content of the specified region using a brief caption. Similar to the image captioning task, metrics like BLEU1, BLEU2, BLEU3, BLEU4, METEOR, ROUGE-L, and SPICE are used to evaluate EarthMarker and other MLLMs, visual promoting models in region target understanding. As displayed in Tab. \ref{tab:Region_caption_compare_dior}, on the DIOR-RSVG test set, compared to other MLLMs such as Qwen-VL-Chat, GeoChat, SPhinx, EarthGPT, and visual prompting models such as Sphinx-V, ViP-LLava, and GLAMM, EarthMarker shows improvements of 7.72$\%$, 9.99$\%$, 11.38$\%$, 12.67$\%$, 12.49$\%$, 7.92$\%$, 10.97$\%$, 165.67$\%$, and 17.50$\%$ in BLEU1, BLEU2, BLEU3, BLEU4, METEOR, ROUGE-L, CIDER and SPICE, respectively. 

\subsection{{Ablation Analysis}}

{\textbf{Visual Encoders.} To investigate the impact of our proposed mixture of visual experts (MOV) strategy on performance, we conduct an ablation experiment, and the results are presented in Tab. \ref{tab:Ablation on visual encoder}. It is clear that the designed MoV achieves the best performance across downstream tasks, compared with solely CNN or ViT\cite{dosovitskiy2021imageworth16x16words} visual backbones. Specifically, when compared to using only CNN or ViT visual backbones, the proposed MoV (CLIP ConvNeXt-L and DINOv2 ViT-L/14) enhances performance in scene classification, region captioning, and referring object classification tasks by 3.88$\%$, 9.41$\%$, and 1.31$\%$, respectively. Furthermore, when compared to the combination of CLIP RN50$\times$16 and DINOv2 ViT-L/14, our designed MoV (CLIP ConvNeXt-L and DINOv2 ViT-L/14) improve performance in scene classification, region captioning, and referring object classification by 2.58$\%$, 6.84$\%$, and 0.43$\%$, respectively. This ablation experiment demonstrates the performance improvement of the MOV module for the proposed EarthMarker.}

{\textbf{Training Data.} We conduct ablation experiments on the training data to investigate the impact of different domains and GPT4V-assisted generated data on the model's final performance. As shown in Tab. \ref{tab:Ablation on training data}, leveraging natural scene captions and referring data significantly enhances the model's performance across various visual tasks, including scene classification, region captioning, and referring object classification. Moreover, incorporating GPT4V-assisted generated data results in the best model performance, particularly showing notable improvements in the region captioning task. Therefore, multi-domain datasets and the long caption data synthesized by GPT4V enable EarthMarker to deliver accurate detailed interpretation of objects in images.}

\subsection{{Computation Analysis and Comparisons}}
{In this part, we compare the GPU memory usage (MiB) and inference time (seconds) of EarthMarker with other MLLMs and visual prompting models in Tab. \ref{tab:Inference Costs}. Specifically, we compare these two aspects of different models on the region captioning task using an NVIDIA RTX A6000. Note that for MLLMs and visual prompting models, GPU memory consumption is primarily driven by storing model weights. As shown in Tab. \ref{tab:Inference Costs}, EarthMarker's GPU memory consumption is larger than that of Sphinx~\cite{lin2024draw}, GeoChat\cite{kuckreja2024geochat}, EarthGPT~\cite{zhang2024earthgpt}, Qwen-VL-Chat~\cite{bai2023qwen}, and Vip-LLava-13b~\cite{cai2024vip}, but lower than that of Sphinx-V~\cite{lin2024draw}. Regarding inference time, similarly, EarthMarker has a longer time consumption than all models except for Sphinx-V, due to its larger model parameters. Fortunately, the difference is mostly within 1 second, so the gap is not significant. In our future work, we will focus on optimizing EarthMarker's inference time, aiming to strike a balance between inference speed and interpretative accuracy. }

\textcolor{red}{
\begin{table}[!h]
\caption{{Computation analysis between our proposed EarthMarker and other MLLMs, visual prompting models.}}\label{tab:Inference Costs}
\renewcommand{\arraystretch}{1.4}
\scalebox{0.84}{
\fontsize{10.3pt}{10.3pt}\selectfont
\begin{tabular}{lcc} %
\toprule
\multicolumn{1}{c}{\!\!\!\! Models\!\!\!\!\!\!\!\!\!\!\!\!}&  \!\! Inference Time(s) &\!\!\!\!\! GPU Memory(MiB) \\ 
\cmidrule(lr){1-1}\cmidrule(lr){2-3}
\textit{\textbf{MLLMs}}          \\ 
\multicolumn{1}{l}{Sphinx}        & 1.08 & 41853              \\
\multicolumn{1}{l}{EarthGPT}        & 0.79    & 41766          \\
\multicolumn{1}{l}{GeoChat}        & 0.74    & 15544          \\
\multicolumn{1}{l}{Qwen-VL-Chat}        & 0.35    & 38145         \\
\cmidrule(lr){1-1}\cmidrule(lr){2-3}
\textit{\textbf{Visual Prompting Models }}          \\ 
\multicolumn{1}{l}{Sphinx-V}             & 1.95   &41794      \\
\multicolumn{1}{l}{Vip-LLava-13B}             & 0.84  &27748          \\
\multicolumn{1}{l}{{\textbf{EarthMarker(Ours)}}}                  {}    & {\textbf{1.63}}  & {\textbf{41038}}          \\ 
\bottomrule
\end{tabular}
} 
\end{table}
}
\begin{figure*}[!t]
	\centering
		\includegraphics[scale=0.152]{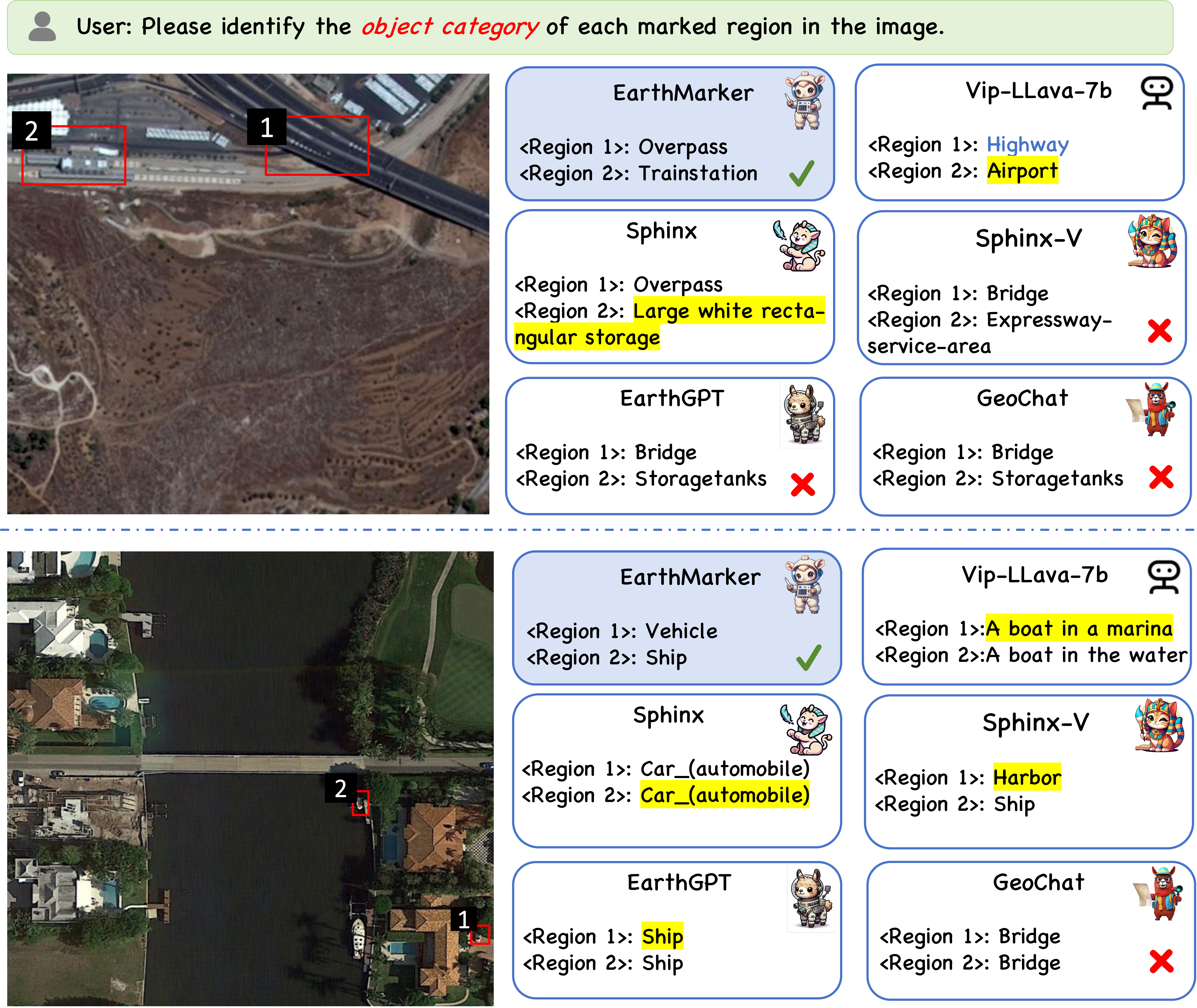}
	      \caption{The referring object classification results on RS images demonstrate superior region-level RS visual understanding capability of EarthMarker compared to other MLLMs and visual prompting models (the symbol $\surd$ indicates consistency with the ground truth, the symbol ${\times}$ represents all incorrect answers, the yellow highlights denote errors, and the blue text represents relatively correct responses.).}
	\label{FIG:Refer}
\end{figure*}

\begin{figure*}[!h]
	\centering
		\includegraphics[scale=0.222]{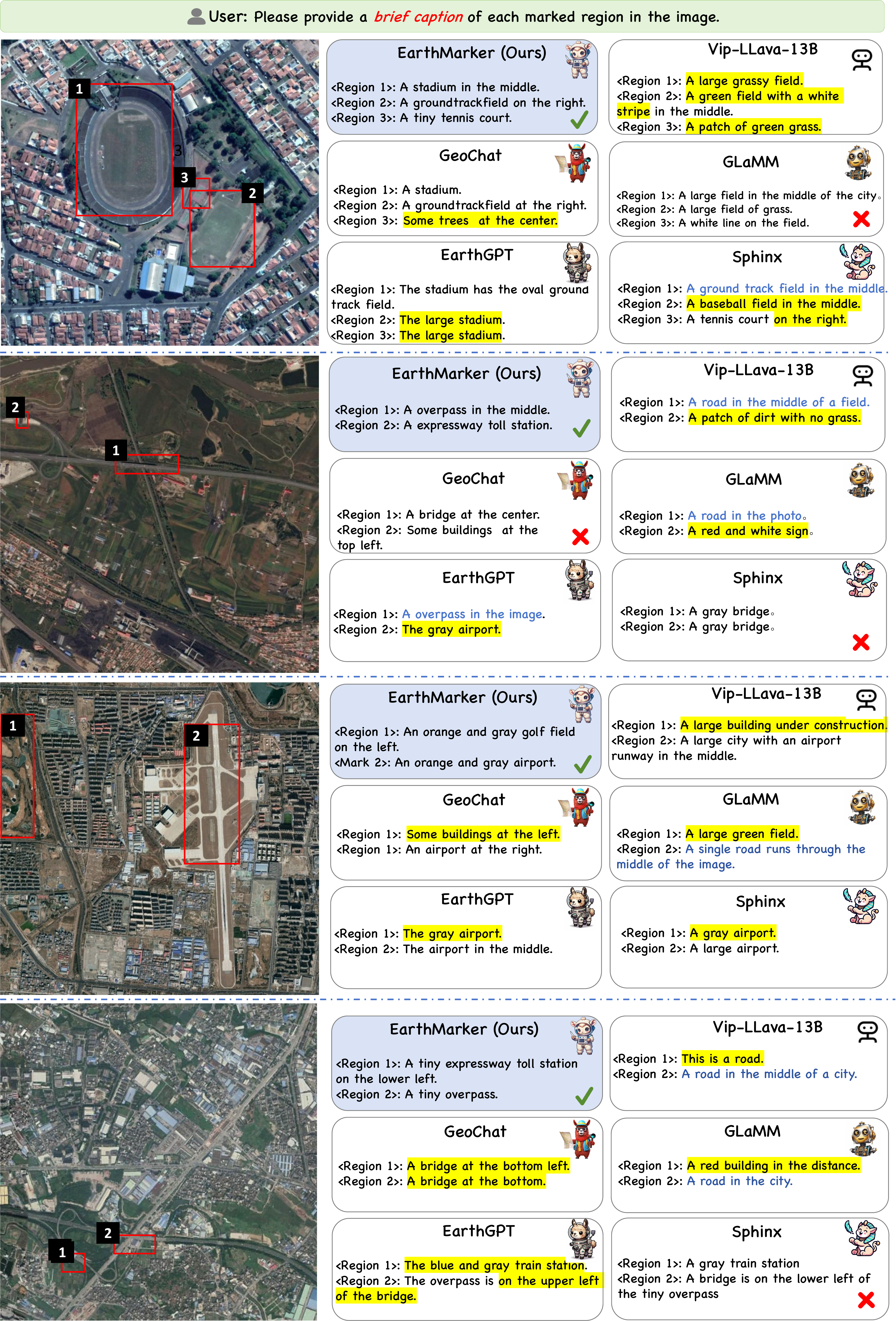}
	      \caption{{The brief region captioning results on RS images demonstrate the precise fine-grained object-level RS visual comprehension ability of EarthMarker, compared to other MLLMs, visual prompting models (the symbol $\surd$ indicates consistency with the ground truth, the symbol ${\times}$ represents all incorrect answers, the yellow highlights denote errors, and the blue text represents relatively correct responses).}}
	\label{FIG:RegionCaption}
\end{figure*}

\begin{figure*}[!t]
	\centering
		\includegraphics[scale=0.2]{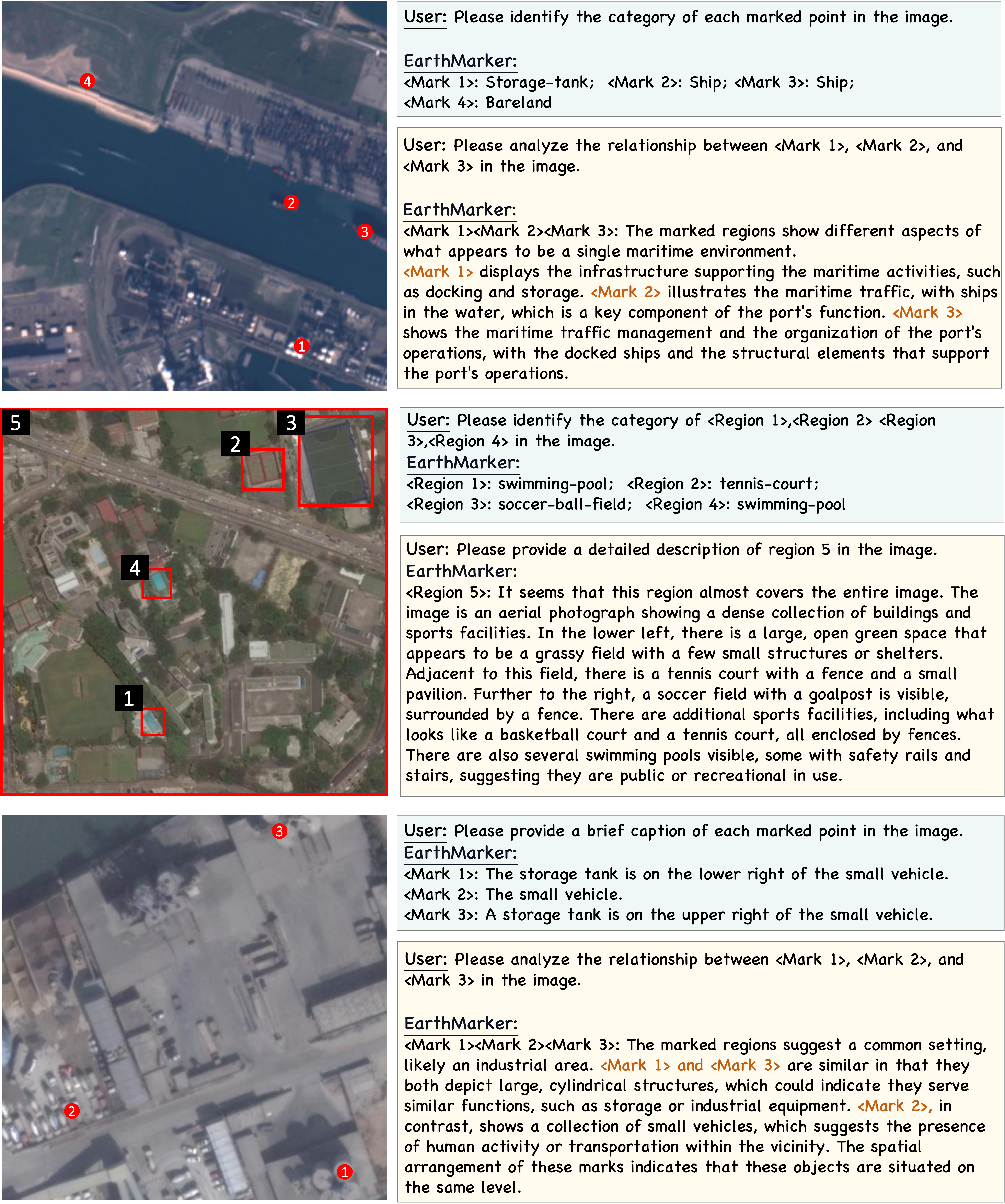}
	      \caption{{Examples of complex low-resolution cases of EarthMarker. EarthMarker can successfully interpret scenes that exhibit challenges such as blurriness, color distortion, and noise interference.}}
	\label{FIG: difficult cases}
\end{figure*}

\begin{figure*}[!t]
	\centering
		\includegraphics[scale=0.14]{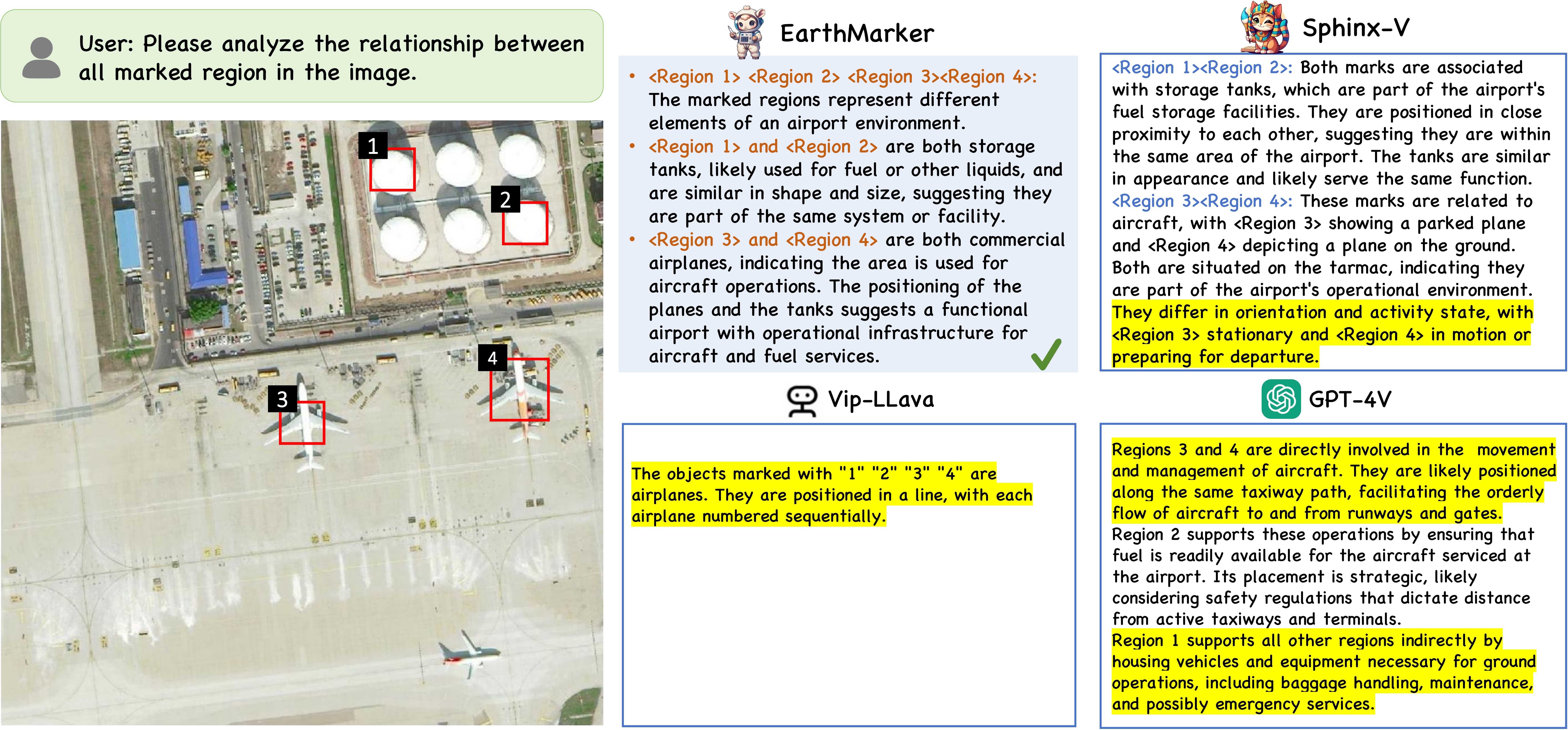}
	      \caption{Examples of completing complex RS tasks such as key object inter-relationship analyses (the yellow highlights denote errors).}
	\label{FIG:gpt}
\end{figure*}
\begin{figure*}[!t]
	\centering 
		\includegraphics[scale=0.14]{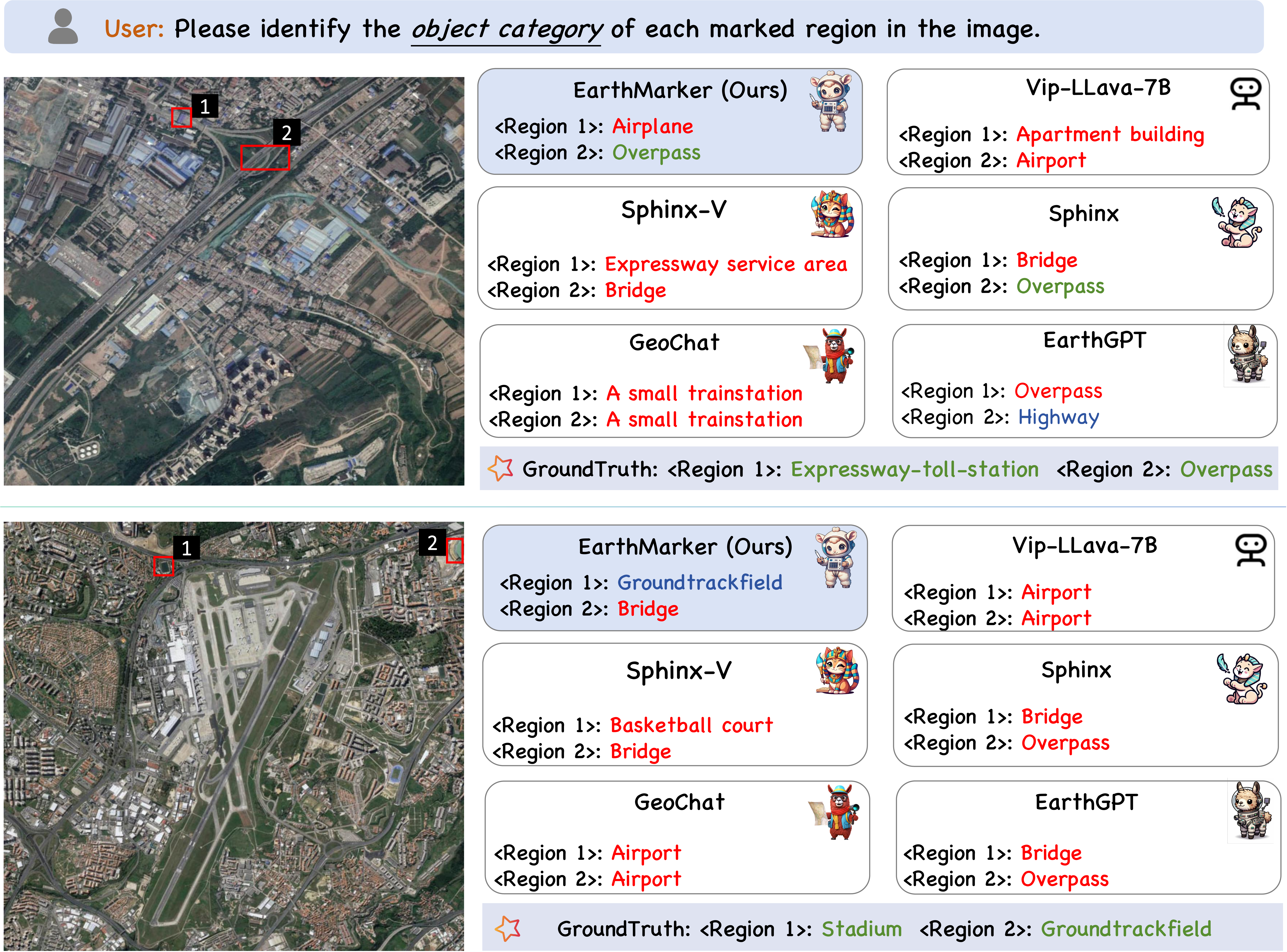}
	      \caption{{Examples of failure cases of EarthMarker (the red highlights denote errors, green highlights denote the right answer and the blue text represents relatively correct responses.)}}
	\label{FIG:failure cases}
\end{figure*}

\section{VISUALIZATION}

\subsection{{Comparison Between EarthMarker and Other Models}}

\subsubsection{{Referring Object Classification}}

In Fig. \ref{FIG:Refer}, there are considerable differences in predictions of regional object categories by EarthMarker and other MLLMs, as well as visual prompting models. {In the top traffic scene image, EarthMarker correctly identifies the ``Overpass" and the ``Train station", whereas other RS-specific MLLMs like GeoChat\cite{kuckreja2024geochat} and EarthGPT\cite{zhang2024earthgpt} misclassify the ``Overpass" as a ``Bridge" and the ``Train Station" as ``Storage Tanks". Other visual prompting models, like Sphinx-V\cite{lin2024draw}, mistakenly identify Region 2 as a ``Expressway-service-area". In the port scene below, only EarthMarker completely interprets the small targets (i.e., car and ship).} It is evident that when faced with complex geographical scenes and blurry tiny objects, EarthMarker's predictions are significantly accurate.

\subsubsection{{Region Captioning}}
We visualize the instances of EarthMarker and other models on the region captioning task, as shown in Fig. \ref{FIG:RegionCaption}. {In complex RS scenarios with numerous, small targets, and extensive geographic coverage, EarthMarker can accurately identify and describe various specified targets (e.g., overpass, tennis court, airport, and stadium). Notably, in the intricate fourth image, EarthMarker can identify small targets, such as expressway toll stations, that are difficult for the human eye to discern. EarthMarker achieves this by integrating information from the surrounding environment, demonstrating an impressive ability to handle both the overall context and detailed local targets. However, interpreting those cases is challenging for other existing models. For example, the current popular MLLMs in the RS domain like GeoChat\cite{kuckreja2024geochat} and EarthGPT\cite{zhang2024earthgpt} are still inferior in regional comprehension of RS imagery compared to our proposed EarthMarker. Natural scene MLLMs Sphinx\cite{lin2023sphinx}, visual prompting models Vip-LLava\cite{cai2024vip} and GLaMM\cite{rasheed2023glamm} struggle to understand complex RS scenarios and interpret those small objects.}

{Thanks to the design of EarthMarker's joint prompting design that integrates both visual and language prompts, and the construction of region-level and point-level visual prompting datasets, EarthMarker achieves exceptional performance in comprehending complicated RS scenarios and analyzing multiple small objects.}

\subsection{{Complex Visual Reasoning}}
\subsubsection{{Low-resolution Scenarios Understanding}}
{The aforementioned cases highlight EarthMarker's strong potential for visual understanding of high-resolution imagery. In this section, we visualize EarthMarker's performance across diverse scenes and varying spatial resolutions. As illustrated in Fig. \ref{FIG: difficult cases}, we present some image cases featuring diverse low resolutions. These images exhibit challenges such as scene blurriness, color distortion, and noise interference, which make manual interpretation relatively difficult. Notably, EarthMarker can still accurately classify objects in diverse scenarios (e.g., foggy harbor environments, urban scenes, and blurred industrial areas), based on the provided region/point-level visual prompts. It can also deliver brief or detailed descriptions of the objects or scenes and analyze the relationships among multiple specified objects. In summary, EarthMarker consistently demonstrates its exceptional capabilities when adapting to new test environments and interpreting scenes with varying spatial resolutions and diverse data sources.}

\subsubsection{Relationship Analyses Tasks}
In this part, we present the qualitative experimental result of EarthMarker to demonstrate its proficiency in completing complex RS tasks such as key object inter-relationship analyses. The text instruction for the relationship analyses task is ``Please analyze the relationship between all marked regions in the image." As shown in Fig. \ref{FIG:gpt}, when faced with an airport scenario, four visual models provided different responses. 

Specifically, the response generated by GPT4V does not specify the exact categories of each marked region. Additionally, GPT4V incorrectly describes Region 1 as ``Region 1 supports all other regions indirectly by housing vehicles and equipment necessary for ground operations, including baggage handling, maintenance, and possibly emergency services", whereas Region 1 actually contains fuel storage tanks. The Vip-LLava model incorrectly identifies all regions as airplanes. Note that Sphinx-V correctly identifies the types of objects in each region and analyzes the internal relationship of some regions. For example, the Sphinx-V answer ``They are positioned in close proximity to each other, suggesting they are within the same area of the airport. The tanks are similar in appearance and likely serve the same function.” However, it fails to provide a comprehensive analysis of the relationships among all four regions, and incorrectly states that the airplane in Region 4 is in motion.

In contrast, our EarthMarker delivers an exemplary response. It firstly summarizes the relationship of all marked regions representing different elements of an airport environment. It then analyzes functionally similar areas in detail, stating that ``$\mathrm{<Region 1>}$ and $\mathrm{<Region 2>}$ are both storage tanks, likely used for fuel or other liquids, and are similar in shape and size, suggesting they are part of the same system or facility. $\mathrm{<Region 3>}$ and $\mathrm{<Region 4>}$ are both commercial airplanes, indicating the area is used for aircraft operations. The positioning of the planes and the tanks suggests a functional airport with operational infrastructure for aircraft and fuel services.” This response accurately reflects the diverse functionalities within the airport, demonstrating superior comprehension and analysis compared to the other visual prompting MLLMs, which plays a  potential key role in solving air-to-ground safety monitoring problems\cite{xu2024collaborative}.

\subsection{{Failure Cases}}
{The aforementioned visualization examples indicate that our proposed EarthMarker possesses strong multi-granularity (e.g., image, region, and point-level) visual reasoning capabilities. Compared to other MLLMs and visual prompting models, EarthMarker demonstrates greater potential for real-world applications. However, in some complex situations and difficult conditions, there are still challenges in accurately interpreting small objects in specific regions. For example, as shown in Fig. \ref{FIG:failure cases}, EarthMarker has limitations in the accuracy of referring object classification for small objects. In the first image, which covers a wide geographic area, we mark two regions: Region 1 (Expressway-Toll-Station), and Region 2 (Overpass). EarthMarker correctly identifies Region 2 but fails to accurately recognize Region 1. In contrast, Sphinx \cite{lin2023sphinx} successfully interpreters  ``Overpass", while EarthGPT provides a close answer, identifying it as a ``Highway". Other models fail to identify the region correctly and exhibit hallucinations. In the second image, all MLLMs \cite{kuckreja2024geochat, zhang2024earthgpt, lin2023sphinx} and visual prompting models\cite{cai2024vip, lin2024draw}, including EarthMarker, struggle to make correct determinations when faced with tiny objects, which occupy only a few pixels in the image. In summary, while the proposed EarthMarker successfully explores the adaptation of visual prompt learning for multi-granularity  RS imagery understanding in the RS domain, there is still significant room for development. In the future, we plan to improve EarthMarker's abilities in visual understanding of tiny objects under challenging scenarios.}

\section{Conclusion}
In this paper, the first visual prompting model called EarthMarker, specifically designed for the RS domain, is proposed. Moreover, the RS visual prompt-based instruction dataset called RSVP is constructed for the first time, containing roughly 3.65 M image-point-text and image-region-text pairings. In addition, the visual prompt learning framework is developed. Particularly, the shared visual encoding method is developed to uniformly refine multi-scale visual features and visual prompt content, which is beneficial for comprehensively understanding the interplay between local visual prompts and the holistic image. Employing the newly construed RSVP and the visual prompt learning framework, EarthMarker is equipped with multi-granularity visual understanding capability across the image, region, and point levels. The proposed EarthMarker advances the development of fine-grained RS imagery comprehension, providing a comprehensive and intelligent analysis in real-world scenarios. {To enhance all-purpose capabilities of EarthMarker, in the future, we plan to incorporate a broader range of various modalities into EarthMarker, enhancing its multi-source, varying spectral imagery and temporal data comprehension capabilities.} Furthermore, we plan to support free-form shapes as visual prompts to adjust the interpretation granularity flexibly.
\bibliographystyle{unsrt}
\bibliography{my.bib}

\end{document}